\documentclass[10pt,twocolumn,letterpaper]{article}

\usepackage{cvpr}              %
\usepackage{xcolor}
\usepackage{tcolorbox}
\usepackage{float}
\usepackage{algorithm}
\usepackage{listings}
\usepackage{titletoc}
\usepackage{minitoc}
\usepackage{booktabs}
\usepackage{multirow}
\usepackage{graphicx}
\usepackage[table]{xcolor}

\floatstyle{ruled}
\restylefloat{algorithm}

\definecolor{pddshade}{RGB}{235,240,243}

\newtcolorbox{pmdkey}{
    colback=pddshade,
    colframe=pddshade,
    boxrule=0pt,
    sharp corners,
    left=6pt,
    right=6pt,
    top=6pt,
    bottom=6pt,
    before skip=7pt,
    after skip=7pt
}

\lstdefinestyle{pmd}{
    language=Python,
    basicstyle=\ttfamily\scriptsize,
    commentstyle=\color{pmdcomment},
    keywordstyle=\color{pmdkeyword}\bfseries,
    emph=[1]{student,teacher},
    emphstyle=[1]\color{pmdmodel}\bfseries,
    emph=[2]{randn,randint,rand_like,randn_like,
              sample_tilt,exp,flatten,chunk,mean,sum,
              detach,alpha,sigma},
    emphstyle=[2]\color{pmdfunction},
    backgroundcolor=\color{pmdcodebg},
    columns=fullflexible,
    keepspaces=true,
    showstringspaces=false,
    breaklines=true,
    breakatwhitespace=true,
    numbers=none,
    frame=none,
    aboveskip=4pt,
    belowskip=4pt,
    xleftmargin=5pt,
    xrightmargin=5pt
}

\definecolor{pmdcodebg}{RGB}{252,244,234}
\definecolor{pmdcomment}{RGB}{170,167,164}
\definecolor{pmdkeyword}{RGB}{81,53,121}
\definecolor{pmdfunction}{RGB}{108,146,171}
\definecolor{pmdmodel}{RGB}{153,44,50}

\usepackage[T1]{fontenc}
\definecolor{cvprblue}{rgb}{0.21,0.49,0.74}
\usepackage[pagebackref,breaklinks,colorlinks,allcolors=cvprblue]{hyperref}

\title{Posterior Matching Distillation}
\author{Hidir Yesiltepe$^1$ \quad
  Julius Berner$^2$ \quad 
  Arash Vahdat$^2$ \quad
  Pinar Yanardag$^1$ \\[0.25em] 
  $^1$Virginia Tech \qquad $^2$NVIDIA \\ [0.5em]
\normalsize{\url{https://pmd-distillation.github.io}}}

\begin{document}
\twocolumn[{
\maketitle
\begin{center}
    \captionsetup{type=figure}
    \vspace{-3.3em}
\newcommand{\imwidth}{1\textwidth}
\begin{tabular}{@{}c@{}}

\parbox{\imwidth}{ \centering 
\vspace{1.5em}
\includegraphics[width=\imwidth, ]{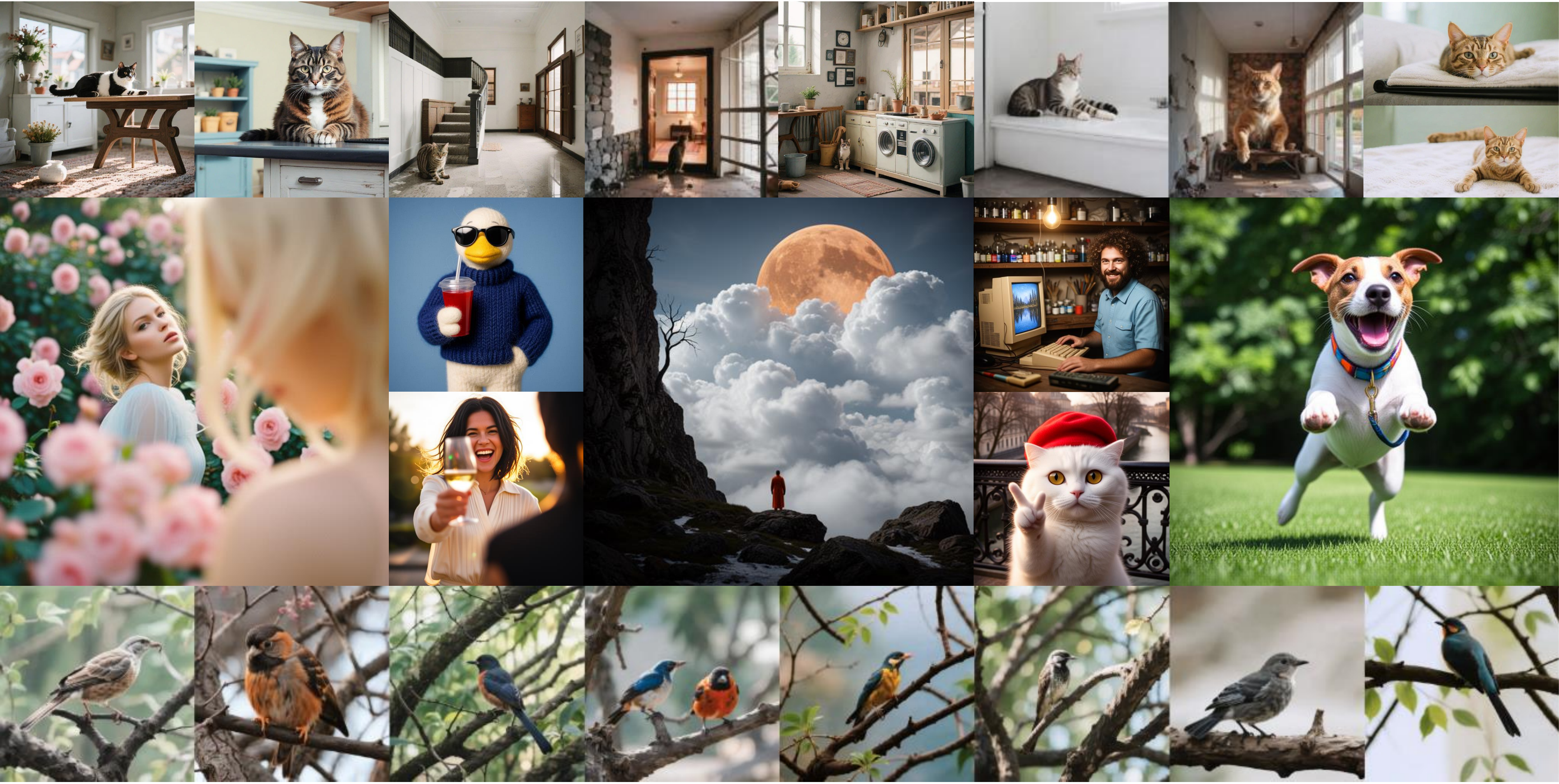}}
\\

\end{tabular}
     \vspace{-0.5em}
    \caption{\textbf{Samples in Four-Step Distillation.} Samples generated by four-step PMD-distilled models. The top and bottom rows show SDXL samples, with each group generated from the same prompt to illustrate conditional diversity across multiple draws. The middle row presents representative samples from FLUX.2 [klein] 4B generated from different prompts, showcasing a variety of visual content.}
    \label{fig:teaser}
\end{center}
}] 

\begin{abstract}
Diffusion distillation accelerates image and video generation by compressing a pretrained teacher's iterative sampling process into a few network evaluations. Distribution-based methods achieve this by matching the student's output distribution to the teacher's. This matching is often driven by reverse-KL objectives, whose mode-seeking bias favors dominant teacher modes at the expense of diversity across samples. We introduce Posterior Matching Distillation (PMD), a new member of the distribution-based distillation family that combines mode-covering posterior sampling with any-step generation. At zero population loss, PMD recovers the teacher's full posterior of clean images at each trained noise level. Its stochastic student supports any-step sampling budgets within the trained timestep grid without further training. To learn these posteriors, PMD matches exponentially reweighted averages of student samples conditioned on a fixed noisy state to corresponding teacher posterior means. Under Gaussian corruption, exponentially tilted posterior mean is a single teacher evaluation at a shifted input. We prove that, matching tilted means over an open neighborhood uniquely identifies the full posterior. PMD thus supervises the entire conditional distribution through a family of mean constraints, without any auxiliary student-score network. Experiments on SDXL and FLUX.2 [klein] 4B demonstrate that PMD achieves the best FID among 4-step methods on both models, while maintaining competitive or leading performance across prompt-alignment and perceptual-quality metrics. 
\end{abstract}

\section{Introduction}
\label{sec:intro}

Diffusion and Flow models \cite{ho2020denoising, song2020score, liu2022flow, lipman2022flow, albergo2025stochastic} have emerged as a powerful approach to generative modeling, particularly for images \cite{podell2024sdxl, esser2024scaling, wu2025qwen, xie2025sana, flux2024, flux-2-2025}  and video \cite{wan2025wan, blattmann2023stable, kong2024hunyuanvideo, yang2025cogvideox, minimax2024video, agarwal2026cosmos}, with growing applications to text \cite{nie2025llada}. Their sampling process progressively transforms noise into a data sample through a sequence of denoising steps, typically requiring many neural network evaluations. Diffusion distillation aims to reduce this computational burden by training a student to generate samples in one or a few steps \cite{salimans2022progressive, song2023consistency, yin2024one, yin2024improved, salimans2024multistep, sabour2026align, boffi2024flow}. However, a noisy state can correspond to many plausible clean samples, while a conventional denoiser provides only a single estimate. This raises a fundamental question: what information about the teacher's generative distribution should a few-step student preserve?

Existing diffusion distillation methods differ in what they seek to preserve from the teacher. Trajectory-based approaches aim to preserve the teacher's sampling process with fewer evaluations. These include progressive distillation \cite{salimans2022progressive}, which compresses successive denoising steps, consistency distillation \cite{song2023consistency}, which learns mappings to trajectory endpoints, and flow-map distillation \cite{boffi2024flow, sabour2026align}, which learns transitions between noise levels. Distribution-based approaches, in contrast, seek to preserve the teacher's generated distribution without requiring the student to follow the same sampling trajectories. Among them, Distribution Matching Distillation (DMD) \cite{yin2024one, yin2024improved} aligns noise-perturbed student and teacher distributions through score-based objectives, while Moment Matching (MM) \cite{salimans2024multistep} enforces agreement between conditional expectations of clean samples at intermediate noisy states.

Matching generated distributions does not necessarily preserve the relationship between noisy inputs and clean samples. A student may reproduce the teacher's overall output distribution while learning a different conditional distribution at each noise level \cite{yin2024one, yin2024improved}. This limitation is particularly relevant to Distribution Matching Distillation, which uses an approximate reverse-KL objective to align noise-perturbed student and teacher distributions. The mode-seeking nature of reverse KL can favor dominant modes at the expense of less frequent outcomes, potentially reducing sample diversity \cite{xu2025one, lu2025adversarial}. Conditional supervision offers an alternative \cite{de2502distributional}. Moment Matching compares the student's and teacher's conditional means, but does so at a new noisy state constructed from the student's prediction, rather than directly at its original input \cite{salimans2024multistep}.

\begin{figure}[t]
    \centering
    \includegraphics[width=1.0\linewidth]{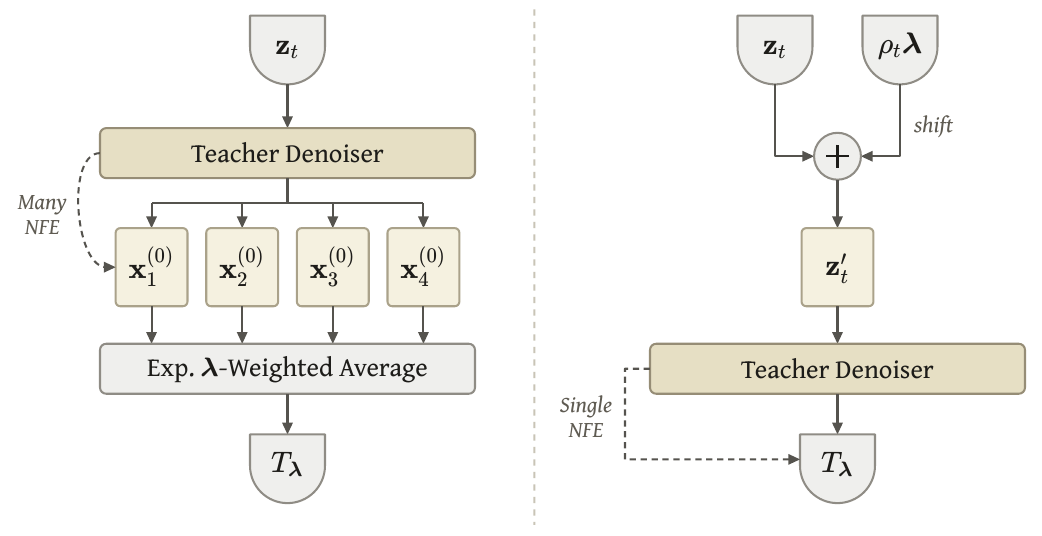} \\
    \vspace{-0.75em}
    \caption{\textbf{Tilt=Shift Identity.} (Left) Estimating a tilted posterior mean requires multiple stochastic reverse-diffusion trajectories from the same noisy state, followed by exponential reweighting of the resulting clean samples. (Right) Under Gaussian corruption, the same mean is obtained with a single frozen-teacher evaluation at a shifted noisy input.}
\label{fig:tilt_equal_shift}
\vspace{-1.1em}
\end{figure}

We introduce Posterior Matching Distillation (PMD), a distribution-based distillation method that trains a stochastic student to recover the teacher's conditional distribution of clean samples at each noise level. PMD matches exponentially reweighted averages of student samples conditioned on a fixed noisy state to corresponding teacher posterior means. In practice, stochasticity is introduced by simply doubling the first layer's input features, and each tilted mean is estimated using as few as four student evaluations per noisy state. The choice of exponential weights is deliberate: under Gaussian corruption, exponentially reweighting the posterior is equivalent to shifting the noisy input at the same noise level \cite{moitra2026steering, holderrieth2025glass, moitra2026tractability, montanari2023sampling, mcdonald2026internal}, an elementary identity we call the \textbf{Tilt=Shift Identity}. As illustrated in Fig.~\ref{fig:tilt_equal_shift}, this identity allows each target mean to be obtained through a single evaluation of the frozen teacher at the shifted input, rather than through computationally expensive Monte Carlo estimation requiring multiple teacher posterior samples. By varying the reweighting direction, PMD imposes a family of constraints that uniquely characterize the teacher's posterior under the stated assumptions.

 PMD's tilted-mean constraints uniquely characterize the teacher's full conditional posterior at each fixed noisy input, under the stated assumptions. Importantly, modes missing from the student's current outputs can remain visible to the objective, providing a learning signal for recovering lost diversity.  Our main contributions are summarized as follows:

\begin{itemize}
    \item \textbf{Direct posterior supervision.} We formulate a conditional distribution-matching objective based on exponentially tilted means. The tilt=shift identity makes these targets accessible through a single frozen-teacher evaluation per tilt, avoiding expensive Monte Carlo sampling from the teacher posterior.

    \item \textbf{Posterior identification and any-step sampling.} We prove that, zero population loss uniquely identifies the full conditional posterior, including its multimodal structure and mode probabilities. An exact posterior student can then reproduce the target distribution without retraining.

    \item \textbf{Superior Conditional Diversity.} Experiments on SDXL and FLUX.2 [klein] 4B demonstrate that PMD consistently outperforms competing methods in intra-prompt diversity across all evaluated metrics, while achieving the best FID among four-step methods.
\end{itemize}

\section{Related Work}
\label{sec:related}

\noindent \textbf{Trajectory-Based Distillation.}
Progressive Distillation \cite{salimans2022progressive} merges consecutive denoising steps, while Consistency Models \cite{song2023consistency}, Latent Consistency Models (LCM) \cite{luo2023latent}, and LCM-LoRA \cite{luo2023lcm} learn mappings to trajectory endpoints. Phased Consistency Models (PCM) \cite{wang2024phased} extend consistency learning to shorter trajectory segments, while SDXL-Lightning \cite{lin2024sdxl} combines progressive and adversarial distillation. Consistency Trajectory Models (CTM) \cite{kim2024consistency} and flow-map methods \cite{boffi2024flow,sabour2026align,geng2026mean,he2026stabilizing} learn transitions between arbitrary time pairs. Recent advances include Parallel Decoding Distillation \cite{shaul2026parallel}, targeting interval-averaged velocities, flexible sampling budgets, and parallel prediction, respectively. On-policy approaches such as \cite{chen2026pi, shaul2026parallel, zhang2026simplifying} imitate teacher velocities along student trajectories , while Simplified Matching Distillation (SMD) \cite{zhang2026simplifying} combines on-policy velocity regression with a training-time residual proxy and repulsion from weaker teacher predictions. 

\noindent \textbf{Distribution-Based Distillation.}
Adversarial Diffusion Distillation (ADD) \cite{sauer2024adversarial} and Latent Adversarial Diffusion Distillation (LADD) \cite{sauer2024fast} use discriminator feedback for few-step generation. Distribution Matching Distillation (DMD) \cite{yin2024one} minimizes an approximate reverse-KL objective using teacher and auxiliary student score estimates, while DMD2 \cite{yin2024improved} improves this framework through two-timescale optimization, adversarial supervision, and multistep backward simulation. More recently, Reward-Tilted Distribution Matching Distillation (RTDMD) \cite{huang2026reinforcing} combines distribution matching with reward-guided reinforcement learning to improve human-preference alignment in few-step generators. Alternative objectives include $f$-divergence matching in f-distill \cite{xu2025one} and Fisher-divergence-based distillation in Score Identity Distillation (SiD) \cite{zhou2024score}. In order to mitigate mode collapse, several works have combined distribution matching with trajectory matching techniques \cite{gu2026anyflow, lin2024sdxl, zheng2026large, yin2024one}. Unlike methods that primarily align generated marginals, PMD directly targets the teacher's full conditional posterior at each noisy input, providing supervision for preserving its multimodal structure.
\begin{figure*}[t]
    \centering
    \includegraphics[width=1.0\linewidth]{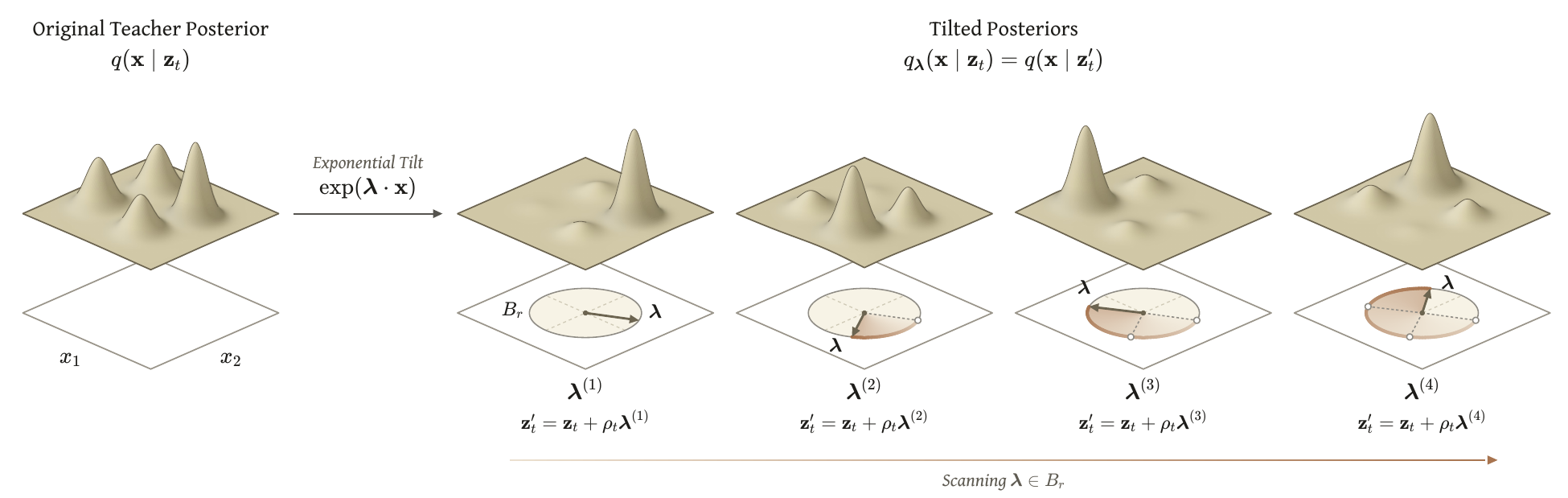} \\
    \vspace{-1em}
    \caption{\textbf{Exponential Tilting of the Teacher Posterior.} Starting from the original teacher posterior \(q(x\mid z_t)\), exponential reweighting by \(\exp(\lambda^\top x)\) changes the relative prominence of its modes. Under Gaussian corruption, each tilted posterior is equivalent to the posterior at a shifted noisy state, \(q_\lambda(x\mid z_t)=q(x\mid z_t+\rho_t\lambda)\). Varying \(\lambda\) over an open neighborhood \(B_r\) emphasizes different modes, producing a family of tilted means that uniquely characterizes the original posterior.}
    \label{fig:posterior_mean_change}
\end{figure*}

\noindent \textbf{Conditional Posterior Modeling.}
DDMs \cite{de2502distributional} train stochastic denoisers to model conditional posteriors using proper scoring rules, while iDDM \cite{martorella2026improved} improves their computational efficiency. Moment Matching (MM) \cite{salimans2024multistep} instead distills a pretrained teacher by matching conditional means at re-noised student predictions, using either an auxiliary denoiser or parameter-space optimization. PMD shares the goal of posterior modeling but obtains supervision directly from a frozen teacher through exponential tilting. 

\noindent \textbf{Exponential Tilting.} Under Gaussian corruption a linear exponential tilt is equivalent to shifting the noisy input at a fixed noise level~\cite{efron2011tweedie,montanari2023sampling}. This connection has been exploited for reward-tilted sampling and alignment~\cite{moitra2026steering,moitra2026tractability} and for reward guidance in discrete diffusion models~\cite{mcdonald2026internal}. Related Gaussian-conditioning identities, in which the shift is accompanied by a change of noise level or covariance, appear in transition sampling and inverse problems~\cite{holderrieth2025glass,mammadov2026exact}. Unlike these works, PMD uses shifted evaluations of a frozen teacher as training targets. It matches a family of tilted posterior means that uniquely identifies the full conditional posterior, with each target obtained from a single teacher evaluation rather than teacher posterior sampling.

\section{Background}
\label{sec:background}

\subsection{Diffusion Models}
\label{sec:background_diffusion}

Let \(x=z_0\sim q_0\) denote a clean image. Diffusion models progressively corrupt \(x\) by adding Gaussian noise, producing a noisy state \(z_t\) at time \(t\in[0,1]\):

\begin{equation}
    z_t=\alpha_t x+\sigma_t \epsilon,
    \qquad \epsilon\sim\mathcal{N}(0,I),
    \label{eq:forward_process}
\end{equation}

\noindent where \(\alpha_t\) and \(\sigma_t\) control the signal and noise strengths, respectively. We adopt the convention that \(t=0\) corresponds to clean data and \(t=1\) to Gaussian noise. This formulation accommodates both conventional diffusion schedules and the linear interpolation used in rectified-flow models.

Let \(q_t(z_t)\) denote the resulting noisy marginal distribution, and \(q(x\mid z_t)\) the conditional distribution of clean images given \(z_t\). An ideal denoising model predicts the conditional mean of this posterior. By Tweedie's formula \cite{efron2011tweedie},

\begin{equation}
    D_t(z_t)
    =\mathbb{E}_{q(x\mid z_t)}[x]
    =\frac{z_t+\sigma_t^2\nabla_{z_t}\log q_t(z_t)}{\alpha_t},
    \label{eq:tweedie}
\end{equation}

\noindent for noise levels where \(\alpha_t>0\) and \(\sigma_t>0\). Thus, the score of the noisy distribution determines the posterior mean. However, this mean is only a single estimate and does not fully characterize the distribution of plausible clean images associated with the same noisy input.

\subsection{Conditional Denoising Posterior}
\label{sec:background_posterior}

The Gaussian corruption process induces a conditional distribution over clean images for each noisy observation \(z_t\). By Bayes' rule, this posterior is given by

\begin{equation}
    q(x\mid z_t)
    \propto q_0(x)
    \exp\left(
        -\frac{\|z_t-\alpha_t x\|^2}{2\sigma_t^2}
    \right),
    \label{eq:denoising_posterior}
\end{equation}

\noindent where \(q_0(x)\) denotes the clean-image distribution. Unlike the posterior mean \(D_t(z_t)\), which provides a single estimate, \(q(x\mid z_t)\) describes the relative probabilities of all plausible clean images associated with the noisy observation. At high noise levels, this posterior may be broad or multimodal, reflecting the ambiguity of recovering a clean image from limited information. Modeling this conditional uncertainty is particularly important for few-step generation, where a single denoising update may need to represent multiple plausible outcomes.

\section{Posterior Matching Distillation}
\label{sec:method}

\subsection{The Tilt=Shift Identity}
\label{sec:tilt_shift}

To obtain information beyond the posterior mean, we consider an exponential tilt of the teacher's conditional distribution. Given a tilt vector $\lambda\in\mathbb{R}^d$, define

\begin{equation}
    q_\lambda(x\mid z_t)
    =
    \frac{
        e^{\lambda^\top x}q(x\mid z_t)
    }{
        \mathbb{E}_{q(x\mid z_t)}
        [e^{\lambda^\top x}]
    }.
    \label{eq:tilted_posterior}
\end{equation}

\noindent The factor $e^{\lambda^\top x}$ changes the relative probabilities of clean images, giving greater weight to samples aligned with $\lambda$, as illustrated in Fig.~\ref{fig:posterior_mean_change}. Different tilts can therefore emphasize different modes of the same posterior. Computing their means would ordinarily require drawing multiple samples from the teacher posterior and reweighting them. However, the Gaussian corruption process admits a much simpler solution.

\begin{pmdkey}
\noindent\textbf{Tilt=Shift Identity.}
For a fixed noise level with $\alpha_t>0$ and $\sigma_t>0$, define $\rho_t=\sigma_t^2/\alpha_t$. Then, for any $\lambda\in\mathbb{R}^d$,

\begin{equation}
    q_\lambda(x\mid z_t)
    =
    q(x\mid z_t+\rho_t\lambda).
    \label{eq:tilt_shift}
\end{equation}

In particular, the mean of the tilted posterior is obtained directly from the teacher denoiser:

\begin{equation}
    \mathbb{E}_{q_\lambda(x\mid z_t)}[x]
    =
    D_t(z_t+\rho_t\lambda).
    \label{eq:tilted_mean}
\end{equation}
\end{pmdkey}

The identity shows that exponentially tilting the posterior is exactly equivalent to shifting the noisy input to the denoiser while keeping the noise level fixed. Consequently, each tilted mean can be obtained through a single evaluation of the frozen teacher at $z_t+\rho_t\lambda$, without sampling from the teacher posterior. The identity holds for any underlying clean-data distribution under Gaussian corruption, assuming an exact posterior-mean denoiser. Its proof is provided in Appendix \ref{app:tilt_shift}.

\begin{figure*}[t]
    \centering
    \includegraphics[width=1.0\linewidth]{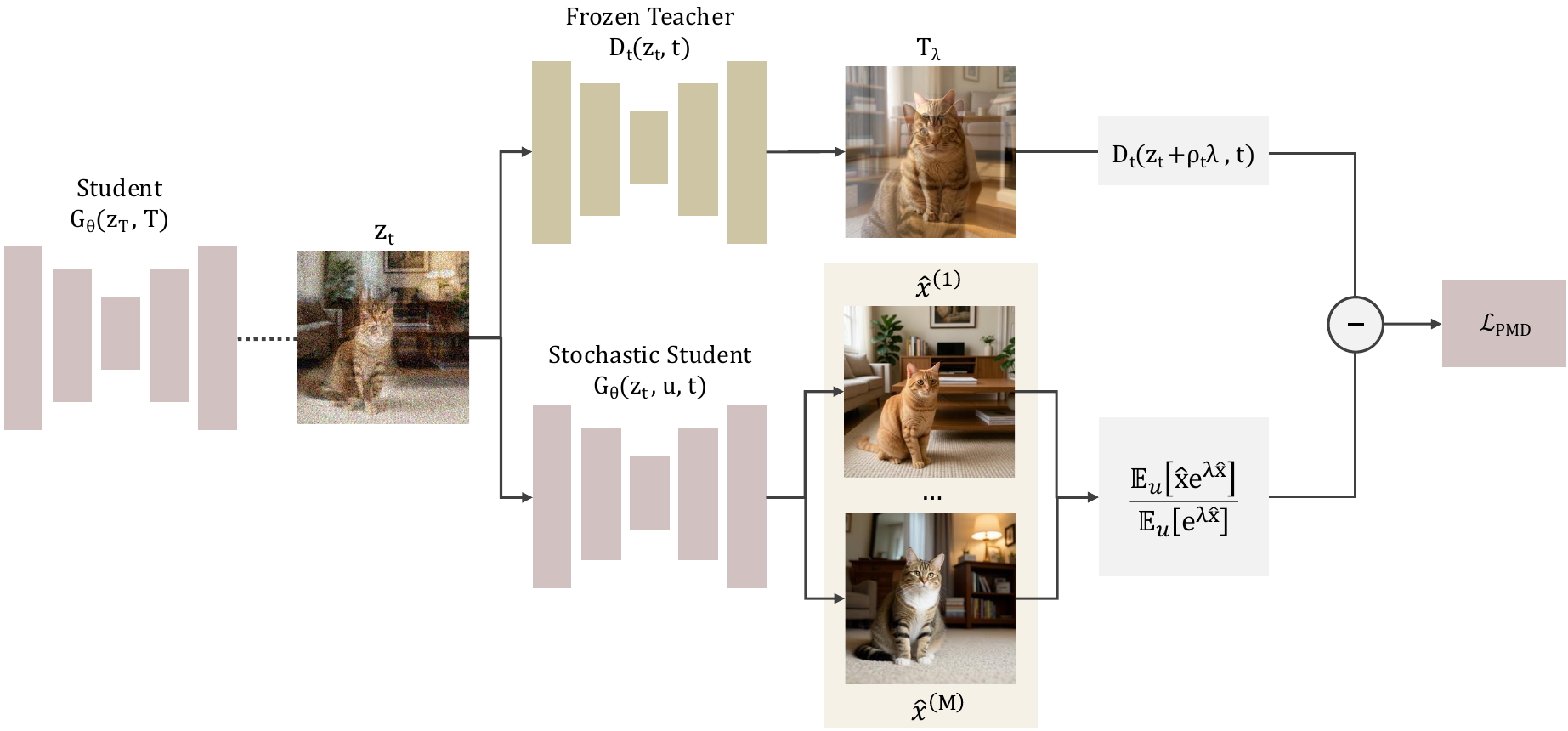} \\
    \vspace{-0.5em}
    \caption{\textbf{PMD Training Framework.}
A noisy state $z_t$ is obtained through the student's on-policy sampling process. The frozen teacher is evaluated at the shifted input $z_t+\rho_t\lambda$, yielding the tilted posterior mean $T_\lambda=D_t(z_t+\rho_t\lambda,t)$. Conditioned on the same $z_t$, the stochastic student generates multiple clean-image predictions  using independent random inputs. These predictions are combined into an exponentially weighted mean,
which is compared with the teacher target $T_\lambda$. PMD minimizes the resulting posterior-matching discrepancy through the loss $\mathcal{L}_{\mathrm{PMD}}$, encouraging the student to recover the teacher's conditional posterior distribution.}
    \label{fig:framework}
\end{figure*}

\subsection{Identifying the Full Conditional Posterior}
\label{sec:posterior_identification}

We now ask whether tilted means contain enough information to determine the entire posterior. Let $G_\theta$ denote a stochastic student that predicts a clean image $\hat{x}=G_\theta(z_t,u,t)$, where $u\sim\mathcal{N}(0,I)$ is an independent random input. Varying $u$ allows the student to produce different clean images from the same noisy state $z_t$, defining a conditional distribution $p_{G_\theta}(x\mid z_t)$.

PMD requires the student's exponentially weighted mean to match the corresponding teacher tilted mean:

\begin{equation}
    \frac{
        \mathbb{E}_u[
            \hat{x}e^{\lambda^\top\hat{x}}
        ]
    }{
        \mathbb{E}_u[
            e^{\lambda^\top\hat{x}}
        ]
    }
    =
    D_t(z_t+\rho_t\lambda),
    \label{eq:pmd_constraint}
\end{equation}

\noindent where the expectations are over $u$, with $z_t$ and $t$ fixed. At $\lambda=0$, this reduces to matching the ordinary posterior mean. Requiring agreement over an open neighborhood of tilts, however, is sufficient to identify the full distribution.

\begin{pmdkey}
\noindent\textbf{Posterior Identification Theorem.}
Suppose the teacher denoiser is exact and the student's exponential moments are finite in an open neighborhood of $\lambda=0$. If Eq.~\eqref{eq:pmd_constraint} holds for every tilt in this neighborhood and almost every noisy input $z_t$ at a fixed noise level, then

\begin{equation}
    p_{G_\theta}(\cdot\mid z_t)=q(\cdot\mid z_t).
    \label{eq:posterior_identification}
\end{equation}

Thus, the student recovers the teacher's entire conditional posterior, not just its mean.
\end{pmdkey}

\noindent The result follows from the fact that tilted means determine the gradient of the log moment-generating function, which uniquely identifies the conditional distribution. Consequently, PMD characterizes the posterior's modes and their probabilities. The full proof is given in Appendix \ref{app:identification}.

\subsection{Posterior Matching Objective}
\label{sec:pmd_objective}

The previous section established that matching tilted means over an open neighborhood uniquely identifies the teacher's conditional posterior. We now use this result to construct a training objective.

For a noisy input $z_t$ and tilt $\lambda$, let $\hat{x}=G_\theta(z_t,u,t)$ denote a stochastic student prediction, where $u\sim\mathcal{N}(0,I)$. The student's exponentially weighted mean is compared directly with the teacher's tilted posterior mean $D_t(z_t+\rho_t\lambda)$. We define the \emph{posterior matching residual} as

\begin{equation}
    r(z_t,\lambda)
    =
    \frac{
        \mathbb{E}_u[
            \hat{x}e^{\lambda^\top\hat{x}}
        ]
    }{
        \mathbb{E}_u[
            e^{\lambda^\top\hat{x}}
        ]
    }
    -
    D_t(z_t+\rho_t\lambda).
    \label{eq:pmd_residual}
\end{equation}

The exponential weights emphasize student predictions aligned with the tilt direction. By the Tilt=Shift Identity, the teacher's corresponding tilted mean is obtained through a single evaluation at the shifted input. Consequently, $r(z_t,\lambda)=0$ if and only if the student's tilted mean matches the teacher's tilted mean in Eq.~\eqref{eq:pmd_constraint}.

\begin{pmdkey}
\noindent\textbf{Posterior Matching Loss.}
PMD minimizes the expected squared posterior matching residual over noise levels, noisy inputs, and tilts:

\begin{equation}
    \mathcal{L}_{\mathrm{PMD}}(\theta)
    =
    \mathbb{E}_{t,z_t,\lambda}
    \left[
        \|r(z_t,\lambda)\|_2^2
    \right].
    \label{eq:pmd_loss}
\end{equation}
\end{pmdkey}

The expectations over stochastic student predictions are taken before computing the squared difference. Thus, the objective matches a property of the student's conditional distribution rather than requiring each individual prediction to equal the teacher's mean. The normalization by the expected exponential weight ensures that the comparison is made directly between the student and teacher tilted means.

\begin{algorithm}[t]
\caption{Posterior Matching Loss}
\label{alg:pmd_loss}
\begin{lstlisting}[style=pmd]
# z, t - noisy input batch and noise level
# M - number of independent student draws

z = z.detach()
tilt = sample_tilt(z, t)

# teacher target via tilt=shift
rho = sigma(t)**2 / alpha(t)
target = teacher(z + rho * tilt, t).detach()

# independent student predictions
x_hat = [student(z, randn_like(z), t)
         for _ in range(M)]

# softmax-weighted student mean
logits = [dot(tilt, x) for x in x_hat]
w = softmax(logits)
student_mean = sum(w[m] * x_hat[m]
                   for m in range(M))

# posterior matching loss
loss = mean(squared_norm(student_mean - target))

return loss
\end{lstlisting}
\end{algorithm}

\subsection{Training PMD}
\label{sec:pmd_training}

We train PMD on noisy states encountered during the student's own sampling process. At each iteration, we select a training noise level $t$ and obtain $z_t$ through backward simulation, starting from Gaussian noise $z_T$. The preceding sampling steps are performed without gradient tracking, and the resulting $z_t$ is detached before computing the loss. 

Given $z_t$, we sample a tilt vector $\lambda$ and evaluate the frozen teacher at the shifted input $z_t+\rho_t\lambda$. By the Tilt=Shift Identity, this single evaluation provides the teacher's tilted posterior mean, $D_t(z_t+\rho_t\lambda)$. We then draw $M$ independent noise inputs $u^{(1)},\ldots,u^{(M)}$ and generate student predictions $\hat{x}^{(m)}=G_\theta(z_t,u^{(m)},t)$. All predictions share the same noisy state and tilt, differing only in their independent random inputs.

We estimate the normalized tilted residual in Eq.~\eqref{eq:pmd_residual} using weighted Monte Carlo samples from the stochastic student. The weights are obtained by applying a softmax to the dot products $\lambda^\top\hat{x}^{(m)}$ between the tilt vector and each student prediction, normalized across the $M$ independent draws. The resulting weighted average estimates the student's tilted mean, which is compared with the teacher's tilted posterior mean using the squared Euclidean distance.

Following DMD2~\cite{yin2024improved} and subsequent distribution-matching distillation methods, we augment the PMD objective with an auxiliary non-saturating logistic GAN loss. A trainable discriminator head operates on caption-conditioned features extracted by the frozen teacher from independently noised reference and student-generated clean latents. The final student objective is
\begin{pmdkey}
\begin{equation}
\mathcal{L}_{\mathrm{student}}
    =
    \mathcal{L}_{\mathrm{PMD}}
    + \gamma_{\mathrm{GAN}}\mathcal{L}_{\mathrm{GAN}}.
\label{eq:pmd_gan}
\end{equation}
\end{pmdkey}

\begin{figure*}[t]
    \centering
    \includegraphics[width=1.0\linewidth]{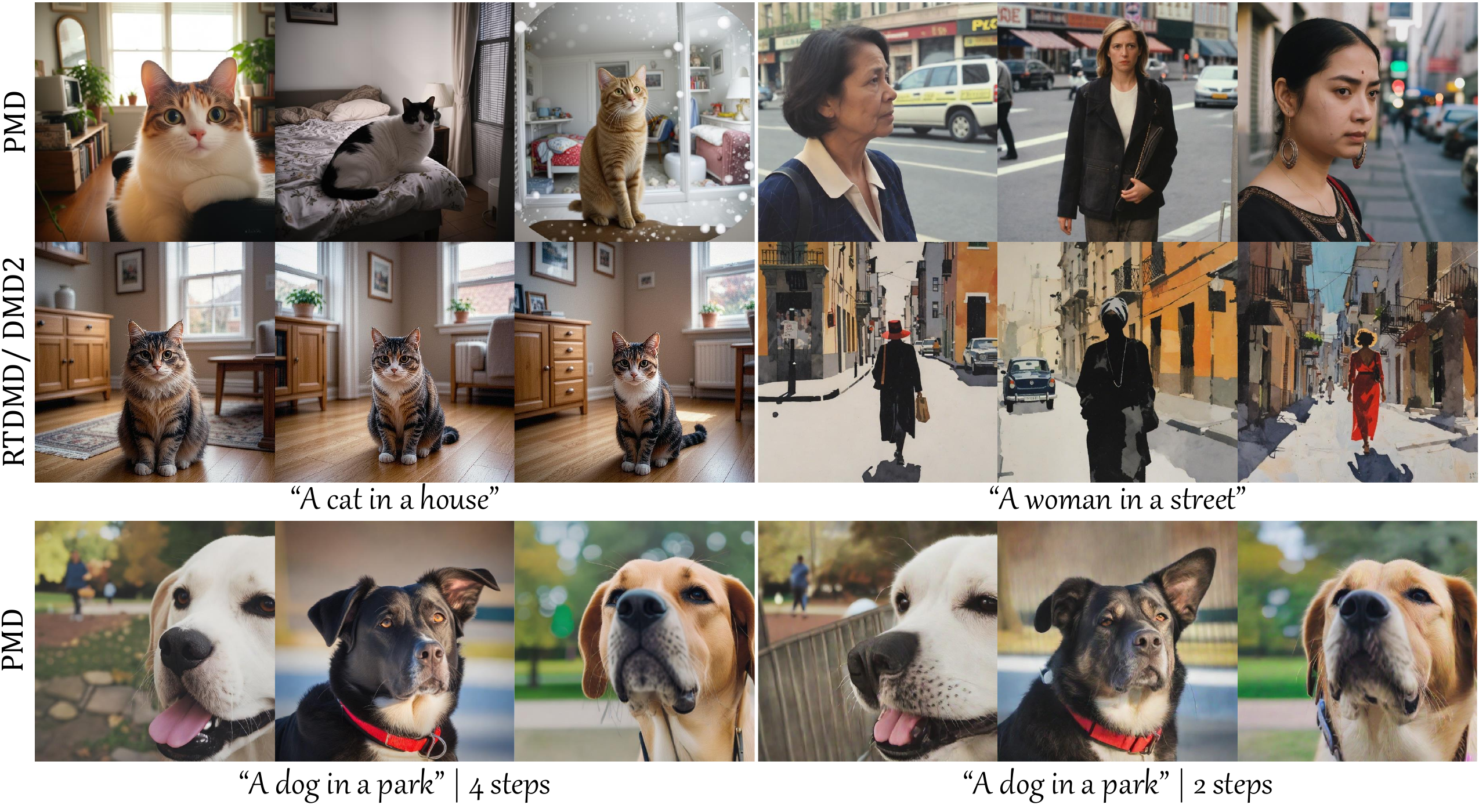} \\
     \vspace{-1.1em}
    \caption{\textbf{Qualitative Comparison and Any-Step Generation.}
PMD demonstrates greater conditional diversity than RTDMD \cite{huang2026reinforcing} on FLUX.2 [klein] 4B (cat) and DMD2 \cite{yin2024improved} on SDXL (woman). The bottom row shows diverse PMD samples generated with both four and two sampling steps using the same model, without retraining.}
    \label{fig:SDXL_PMD_DMD2_comparison}
\end{figure*}

\begin{table*}[t]
\centering
\small
\setlength{\tabcolsep}{4pt}
\resizebox{\textwidth}{!}{%
\begin{tabular}{@{}lcccccccc@{}}
\toprule
Method & NFE & GenEval$\uparrow$ & DPG-Bench$\uparrow$ & OneIG-EN$\uparrow$ & HPSv3$\uparrow$ & PickScore$\uparrow$ & CLIP-S$\uparrow$ & FID$\downarrow$ \\
\midrule

\rowcolor{orange!10}[0pt][0pt]
\multicolumn{9}{c}{\textbf{SDXL}} \\
\midrule

\textcolor{gray}{Euler}
& \textcolor{gray}{$50\times2$}
& \textcolor{gray}{0.5451} & \textcolor{gray}{73.64} & \textcolor{gray}{0.3213}
& \textcolor{gray}{8.23} & \textcolor{gray}{22.28}
& \textcolor{gray}{33.81} & \textcolor{gray}{18.24} \\
\midrule

LCM-LoRA \cite{luo2023lcm}
    & 4
    & 0.5023 & 69.47 & 0.2966
    & 6.58 & 22.05 & 32.59 & 21.83 \\

SDXL-Turbo \cite{sauer2024adversarial}
    & 4
    & 0.4961 & 73.47 & 0.2843
    & 8.99 & 22.18 & 32.87 & 25.91 \\

SDXL-Lightning \cite{lin2024sdxl}
    & 4
    & \underline{0.5564} & 70.99 & 0.3105
    & 9.66 & 22.65 & \underline{33.16} & 24.88 \\

Flash \cite{chadebec2025flash}
    & 4
    & 0.4917 & 68.72 & 0.3089
    & 7.35 & 22.06 & 32.94 & 21.56 \\

PCM \cite{wang2024phased}
    & 4
    & 0.5331 & 68.65 & 0.3031
    & 9.15 & 22.42 & 32.47 & 27.02 \\

DMD2 \cite{yin2024improved}
    & 4
    & \textbf{0.5882} & \textbf{74.49} & \underline{0.3218}
    & \textbf{10.35} & \underline{22.82}
    & \textbf{33.87} & \underline{19.20} \\

Moment Matching \cite{salimans2024multistep}
    & 4
    & 0.3981 & 65.95 & 0.3026
    & 8.02 & 21.81 & 30.70 & 21.61 \\

\textbf{PMD (Ours)}
    & 4
    & 0.5539 & \underline{73.95} & \textbf{0.3426}
    & \underline{10.02} & \textbf{22.97}
    & 32.70 & \textbf{19.14} \\

\addlinespace[4pt]
\midrule
\rowcolor{orange!10}[0pt][0pt]
\multicolumn{9}{c}{\textbf{FLUX.2 [klein] 4B}} \\
\midrule

\textcolor{gray}{Euler}
& \textcolor{gray}{$50\times2$}
& \textcolor{gray}{0.7733} & \textcolor{gray}{83.99} & \textcolor{gray}{0.4981}
& \textcolor{gray}{11.67} & \textcolor{gray}{22.52}
& \textcolor{gray}{33.72} & \textcolor{gray}{21.54} \\
\midrule

FLUX.2 [klein] 4B \cite{flux-2-2025}
    & 4
    & \textbf{0.8259} & \textbf{85.61} & \textbf{0.5098}
    & \underline{12.41} & \underline{22.85}
    & \underline{33.79} & \underline{24.58} \\

SMD \cite{zhang2026simplifying}
    & 4
    & 0.7945 & 83.19 & 0.4724
    & 11.61 & 22.59
    & 33.72 & 25.89 \\

Moment Matching \cite{salimans2024multistep}
    & 4
    & 0.7169 & 80.02 & 0.4310
    & 12.18 & 22.33
    & 33.03 & 27.19 \\

\textbf{PMD (Ours)}
    & 4
    & \underline{0.8163} & \underline{84.25} & \underline{0.4936}
    & \textbf{12.58} & \textbf{22.97}
    & \textbf{34.13} & \textbf{24.12} \\

\bottomrule
\end{tabular}%
}
\vspace{-0.75em}

\caption{\textbf{Quantitative Comparison of Image Generation Methods.}
Performance comparison on SDXL and FLUX.2 [klein] 4B at $1024 \times 1024$ resolution. HPSv3, PickScore, CLIP-S, and FID are evaluated using 10K prompts from MS-COCO 2014 \cite{lin2014microsoft}. Best and second-best results among four-step methods are highlighted in bold and underlined, respectively.}
\label{tab:sdxl_flux_benchmarks}
\end{table*}

\subsection{Any-Step Sampling}
\label{sec:any_step_sampling}

PMD supports flexible sampling budgets without training separate students for different numbers of steps. Unlike flow-map methods that explicitly learn transitions between pairs of noise levels, PMD learns the conditional posterior of clean images at each noise level. During sampling, the student predicts a clean image from the current noisy state. If another step is required, this prediction is re-noised to the next selected level using the Gaussian corruption schedule, and the student is evaluated again. The final prediction is returned without re-noising.

\section{Experiments}
\label{sec:experiments}

\begin{figure*}[t]
    \centering
    \includegraphics[width=1.0\linewidth]{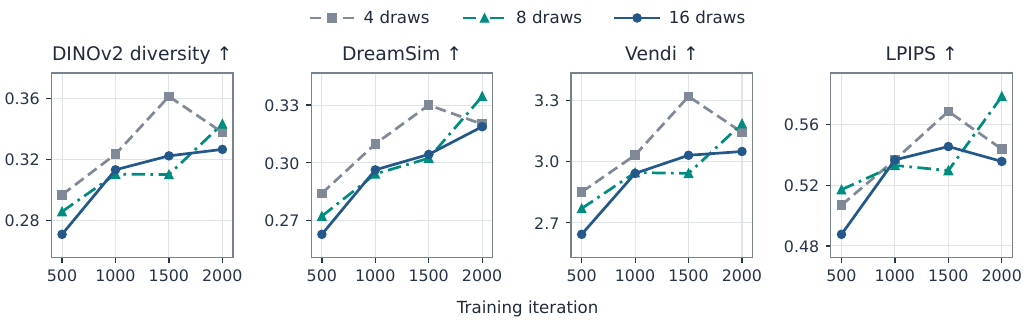} \\
    \vspace{-1.1em}
    \caption{\textbf{Ablation on the Number of Student Draws.}
Diversity scores during four-step PMD distillation of FLUX.2 [klein] 4B, using 4, 8, or 16 independent student draws per noisy input. We evaluate sample diversity using DINOv2, DreamSim, Vendi, and LPIPS.}
    \label{fig:ablation_draws}
\vspace{-1.5em}
\end{figure*}

\begin{figure}[t]
    \centering
    \includegraphics[width=1.0\linewidth]{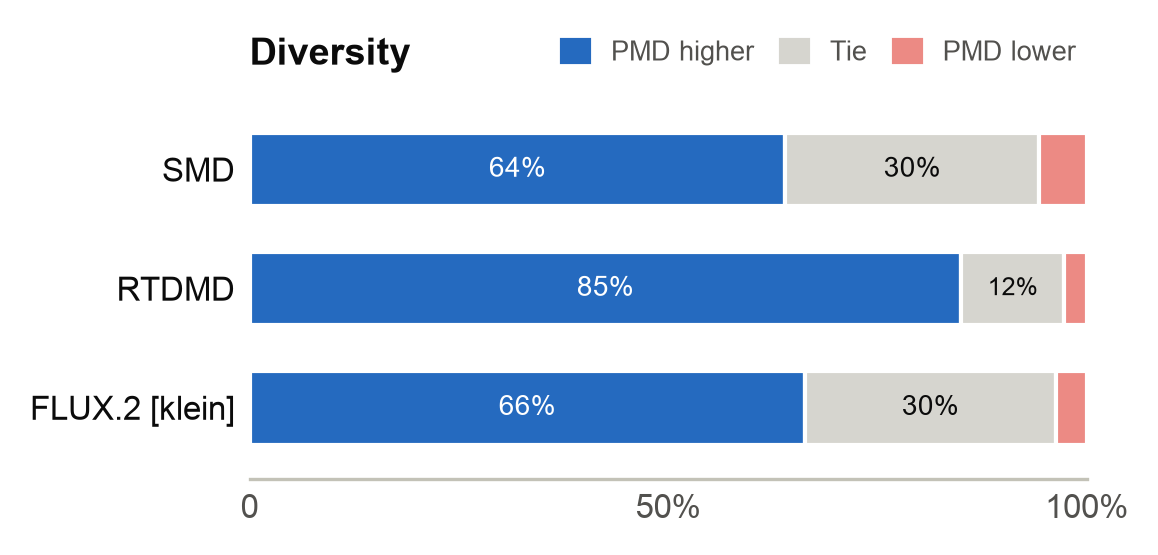} \\
    \caption{\textbf{User Study for FLUX.2 [klein] 4B four-step models.}}
    \label{fig:diversity_user_study}
\end{figure}

\begin{table}[t]
\centering
\caption{
    \textbf{Diversity comparison on PartiPrompts.}
    Results are averaged over 1,632 prompts with 8 seeds
    per prompt using 4-steps.
    Higher values indicate greater diversity across all metrics.
}
\vspace{-0.75em}
\label{tab:diversity}
\setlength{\tabcolsep}{3pt}
\renewcommand{\arraystretch}{1.1}
\resizebox{\columnwidth}{!}{%
\begin{tabular}{lcccc}
\toprule
Method
    & Vendi $\uparrow$
    & LPIPS $\uparrow$
    & DINOv2 $\uparrow$
    & DreamSim $\uparrow$ \\
\midrule
LCM-LoRA
    & 3.1186 & 0.5814 & 0.3391 & 0.3101 \\
SDXL-Turbo
    & 2.4781 & 0.5569 & 0.2470 & 0.2181 \\
SDXL-Lightning
    & 2.8656 & 0.5749 & 0.3050 & 0.2847 \\
Flash
    & \underline{3.2198} & 0.5912 & \underline{0.3484} & \underline{0.3153} \\
PCM
    & 2.9537 & \underline{0.5922} & 0.3204 & 0.2937 \\
DMD2
    & 2.5178 & 0.5715 & 0.2523 & 0.2406 \\
\midrule
\rowcolor{orange!10}
\textbf{PMD (SDXL)}
    & \textbf{4.4621}
    & \textbf{0.6329}
    & \textbf{0.5178}
    & \textbf{0.4599} \\
\bottomrule
FLUX.2 [klein] 4B
    & 2.3237 & 0.4459 & 0.2171 & 0.1963 \\
SMD
    & \underline{2.4104} & \underline{0.5031} & 0.2648 & 0.2487 \\
RTDMD
    & 1.8540 & 0.3729 & 0.1448 & 0.1168 \\
Moment Matching
    & 2.3891 & 0.4995 & \underline{0.2898} & \underline{0.2507} \\
\midrule
\rowcolor{orange!10}
\textbf{PMD (FLUX.2)}
    & \textbf{3.3708} & \textbf{0.5705} & \textbf{0.3690} & \textbf{0.3503} \\
\bottomrule
\end{tabular}%
}
\vspace{-1em}
\end{table}

\paragraph{Models and baselines.}
We evaluate Posterior Matching Distillation (PMD) on
text-to-image models from two families: SDXL \cite{podell2024sdxl} and
FLUX.2~[klein] 4B \cite{flux-2-2025}. Our primary setting uses four sampling steps. For SDXL, we compare against LCM-LoRA \cite{luo2023lcm}, SDXL-Turbo \cite{sauer2024adversarial}, SDXL-Lightning \cite{lin2024sdxl}, Flash \cite{chadebec2025flash}, PCM \cite{wang2024phased}, DMD2 \cite{yin2024improved}, and Moment Matching \cite{salimans2024multistep}, as well as an undistilled teacher reference. For FLUX.2~[klein] 4B, the comparison includes the base model, SMD \cite{zhang2026simplifying}, RTDMD \cite{huang2026reinforcing}, and Moment Matching \cite{salimans2024multistep}. The number of network function evaluations (NFE) is reported alongside generation quality to make the inference budget explicit.

\noindent \textbf{Implementation details.}
All experiments are conducted on 8 NVIDIA L40 GPUs for 2000 student updates, with a batch size of 16. We use a discriminator-to-generator update ratio of 5:1. We use 4 student draws per training state. We optimize with AdamW ($\beta_1=0.9$, $\beta_2=0.999$), a learning rate of $5\times10^{-7}$, gradient norm clipping at 10, and a fixed GAN loss weight of $0.75$.
We initialize the student from the undistilled pretrained model. To enable stochasticity, we concatenate Gaussian noise-mixed and original latent tokens, doubling the input projection dimension. Further architectural details are provided in the Appendix \ref{sec:appendix_implementation_details}.

\noindent \textbf{Evaluation metrics.}
We measure generation quality and prompt adherence using GenEval \cite{ghosh2023geneval}, DPG-Bench \cite{hu2024ella}, OneIG-EN \cite{chang2026oneig}, HPSv3 \cite{ma2025hpsv3}, PickScore \cite{kirstain2023pick}, and CLIP \cite{radford2021learning} Score (CLIP-S), and report FID \cite{heusel2017gans} as an additional distributional quality measure. To assess intra-prompt diversity independently of prompt adherence, we use the Vendi score \cite{friedman2022vendi} and perceptual/feature-space diversity metrics based on LPIPS \cite{zhang2018unreasonable}, DINOv2 \cite{oquab2023dinov2}, and DreamSim \cite{fu2023dreamsim}. Metric details are shared in Appendix \ref{app:eval_metrics}.

\noindent \textbf{Qualitative Results.} Fig.~\ref{fig:SDXL_PMD_DMD2_comparison} provides qualitative comparisons of PMD against DMD2 on SDXL and RTDMD on FLUX.2 [klein] 4B under identical text prompts. Across both model families, PMD exhibits greater variation in subject appearance, scene composition, and backgrounds while preserving the semantic content of the prompts. The figure also demonstrates PMD's any-step capability: the same trained student generates diverse samples with four and two inference steps. Additional results are provided in Appendix~\ref{sec:Flux_results}.

\noindent \textbf{Quantitative Results.} Table~\ref{tab:sdxl_flux_benchmarks} shows that PMD achieves the best FID among the compared four-step methods on both SDXL and FLUX.2 [klein] 4B, while leading in OneIG-EN and PickScore on SDXL and HPSv3, PickScore, and CLIP-S on FLUX.2. RTDMD \cite{huang2026reinforcing}, which uses additional external reward supervision, achieves GenEval, HPSv3, and PickScore scores of 0.9076, 15.59, and 23.99 on FLUX.2, but a substantially worse FID of 31.57 compared to PMD's 24.12. User Studies with 50 users are shared in Fig.~\ref{fig:diversity_user_study}  and Appendix~\ref{app:user_study}.

\noindent \textbf{Intra-prompt diversity.}
We evaluate conditional diversity on PartiPrompts \cite{yu2022scaling} using four complementary metrics in Table~\ref{tab:diversity}. PMD consistently outperforms all baselines across both model families. On SDXL, PMD improves over the strongest baseline for each metric by 48.6\% in DINOv2, 45.9\% in DreamSim, 38.6\% in Vendi, and 6.9\% in LPIPS. On FLUX.2 [klein] 4B, the corresponding improvements are 27.3\%, 39.7\%, 39.8\%, and 13.4\%, respectively. These consistent gains demonstrate PMD's ability to preserve diverse generations under fixed noise, and text prompts.

\begin{figure}[t]
    \centering
    \includegraphics[width=1.0\linewidth]{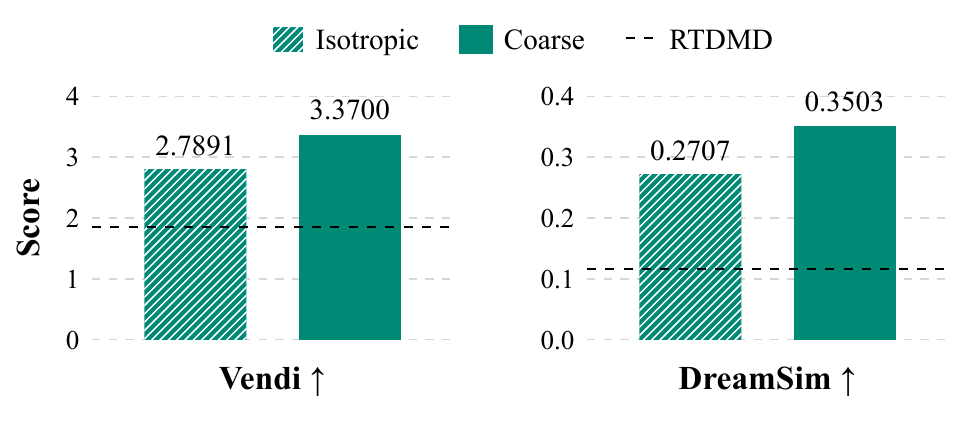} 
    \caption{\textbf{Ablation on Tilt Selection.}} 
    \label{fig:ablation_tilt_selection}
    \vspace{-15pt}
\end{figure}

\noindent \textbf{Ablation on tilt directions.}
We investigate two tilt-sampling strategies: isotropic, which samples independent Gaussian directions at each spatial and channel position, and coarse, which samples directions at a lower spatial resolution and upsamples them to the latent resolution, introducing spatial correlations. As shown in Fig.~\ref{fig:ablation_tilt_selection}, both strategies achieve strong diversity, with coarse sampling consistently yielding better results. Details are provided in Appendix~\ref{sec:analysis_tilt_direction}. 

\noindent \textbf{Ablation on stochastic student draws.}
We examine the Monte Carlo
budget using $\{4,8,16\}$ stochastic student draws.
Fig.~\ref{fig:ablation_draws} plots DINOv2
diversity, DreamSim diversity, Vendi score,
and LPIPS at training iterations. More ablations are provided in Appendix \ref{app:more_ablations}.

\vspace{-5pt}
\section{Conclusion and Limitation}
We introduced Posterior Matching Distillation (PMD), a distillation method that directly targets the teacher's conditional posterior through exponentially tilted means obtained from single frozen-teacher evaluations. We prove that these constraints uniquely identify the full posterior. Experiments demonstrate the best diversity. While PMD provides direct posterior supervision, it requires multiple student draws during training and still relies on an auxiliary adversarial loss to further improve visual realism.

{
    \small
    \bibliographystyle{ieeenat_fullname}
    \bibliography{main}
}
\clearpage
\clearpage
\appendix
\setcounter{page}{1}
\maketitlesupplementary

\section*{Table of Contents}
\addcontentsline{toc}{section}{Supplementary Material Table of Contents}
\startcontents[appendix]
\printcontents[appendix]{l}{1}{\setcounter{tocdepth}{2}}

\section{Proofs}
\label{app:proofs}

\subsection{Proof of Tilt=Shift Identity}
\label{app:tilt_shift}

\noindent\textbf{Setup.} Let $x\sim q_0$ denote a clean image, and let $z_t=\alpha_t x+\sigma_t\epsilon$ with $\epsilon\sim\mathcal{N}(0,I)$, where $\alpha_t>0$ and $\sigma_t>0$. By Bayes' rule, the posterior of $x$ given $z_t$ is

\begin{equation}
  q(x \mid z_t) \propto q_0(x)
  \exp\!\left(-\frac{\|z_t-\alpha_t x\|^2}{2\sigma_t^2}\right),
  \label{eq:appA_posterior}
\end{equation}

\noindent and the denoiser $D_t(z_t)=\mathbb{E}_{q(x\mid z_t)}[x]$ is its mean. For a tilt vector $\lambda\in\mathbb{R}^d$, the tilted posterior is

\begin{equation}
  q_\lambda(x \mid z_t)
  = \frac{e^{\lambda^\top x}q(x \mid z_t)}
  {\mathbb{E}_{q(x\mid z_t)}[e^{\lambda^\top x}]}.
  \label{eq:appA_tilt}
\end{equation}

\begin{pmdkey}
\noindent\textbf{Tilt=Shift Identity.}
Let $\rho_t=\sigma_t^2/\alpha_t$. Then, for any $\lambda\in\mathbb{R}^d$,

\begin{equation}
  q_\lambda(x \mid z_t)
  = q(x \mid z_t+\rho_t\lambda).
  \label{eq:appA_shift}
\end{equation}

In particular, the tilted posterior mean is

\begin{equation}
  \mathbb{E}_{q_\lambda(x\mid z_t)}[x]
  = D_t(z_t+\rho_t\lambda).
  \label{eq:appA_mean}
\end{equation}
\end{pmdkey}

\noindent\textit{Proof.}
Expanding the square in Eq.~\eqref{eq:appA_posterior} gives

\begin{equation*}
  \frac{\|z_t-\alpha_t x\|^2}{2\sigma_t^2}
  = \frac{\|z_t\|^2}{2\sigma_t^2}
  - \frac{z_t^\top x}{\rho_t}
  + \frac{\alpha_t^2\|x\|^2}{2\sigma_t^2},
\end{equation*}

\noindent where we used $\alpha_t/\sigma_t^2=1/\rho_t$. The first term does not depend on $x$ and is absorbed into the normalizing constant, so

\begin{equation}
  q(x \mid z_t) \propto q_0(x)
  \exp\!\left(-\frac{\alpha_t^2\|x\|^2}{2\sigma_t^2}\right)
  \exp\!\left(\frac{z_t^\top x}{\rho_t}\right).
  \label{eq:appA_expanded}
\end{equation}

Thus, the noisy state $z_t$ enters the posterior only through the last factor. Since $e^{\lambda^\top x}=\exp\big((\rho_t\lambda)^\top x/\rho_t\big)$, multiplying Eq.~\eqref{eq:appA_expanded} by $e^{\lambda^\top x}$ simply shifts $z_t$ to $z_t+\rho_t\lambda$:

\begin{equation}
  \begin{split}
    e^{\lambda^\top x} q(x \mid z_t)
    &\propto q_0(x)
    \exp\!\left(-\frac{\alpha_t^2\|x\|^2}{2\sigma_t^2}\right)\\
    &\qquad\times
    \exp\!\left(\frac{(z_t+\rho_t\lambda)^\top x}{\rho_t}\right).
  \end{split}
  \label{eq:appA_tilted}
\end{equation}

\noindent The right-hand side of Eq.~\eqref{eq:appA_tilted} is the right-hand side of Eq.~\eqref{eq:appA_expanded} evaluated at the noisy state $z_t+\rho_t\lambda$, so it is proportional to $q(x \mid z_t+\rho_t\lambda)$. It is also integrable in $x$, being $q_0(x)$ times a function that is bounded in $x$ (the quadratic term in the exponent dominates the linear one); hence the normalizer $\mathbb{E}_{q(x\mid z_t)}[e^{\lambda^\top x}]$ in Eq.~\eqref{eq:appA_tilt} is finite for every $\lambda$.

Normalizing both sides of Eq.~\eqref{eq:appA_tilted} over $x$, the left-hand side becomes $q_\lambda(x \mid z_t)$ by Eq.~\eqref{eq:appA_tilt} and the right-hand side becomes $q(x \mid z_t+\rho_t\lambda)$, which proves Eq.~\eqref{eq:appA_shift}.

Taking the mean of $x$ on both sides of Eq.~\eqref{eq:appA_shift} and using the definition of the denoiser at the shifted state $z_t+\rho_t\lambda$,

\begin{equation*}
  \mathbb{E}_{q_\lambda(x\mid z_t)}[x]
  = \mathbb{E}_{q(x\mid z_t+\rho_t\lambda)}[x]
  = D_t(z_t+\rho_t\lambda),
\end{equation*}

\noindent which proves Eq.~\eqref{eq:appA_mean}. The argument uses only the Gaussian form of Eq.~\eqref{eq:appA_posterior}, so the identity holds for any clean-image distribution $q_0$. \hfill$\square$

\subsection{Proof of Posterior Identification Theorem}
\label{app:identification}

\noindent\textbf{Setup.}
Fix a noise level with $\alpha_t>0$ and $\sigma_t>0$, and let $\rho_t=\sigma_t^2/\alpha_t$. For a clean image $x=z_0\sim q_0$ and noisy state $z_t=\alpha_t x+\sigma_t\epsilon$ with $\epsilon\sim\mathcal{N}(0,I)$, the teacher posterior is

\begin{equation}
  q(x \mid z_t) \propto q_0(x)
  \exp\!\left(-\frac{\|z_t-\alpha_t x\|^2}{2\sigma_t^2}\right),
  \label{eq:appB_posterior}
\end{equation}

and an exact teacher denoiser returns its mean, $D_t(z_t)=\mathbb{E}_{q(x\mid z_t)}[x]$. The stochastic student predicts $\hat{x}=G_\theta(z_t,u,t)$, where $u\sim\mathcal{N}(0,I)$ is an independent random input. We denote the conditional distribution of $\hat{x}$ by $p_{G_\theta}(x\mid z_t)$, with expectations $\mathbb{E}_u[\cdot]$ taken over $u$ for fixed $z_t$ and $t$. Let $B_r=\{\lambda\in\mathbb{R}^d:\|\lambda\|<r\}$, with $r>0$, be an open ball centered at $\lambda=0$.

\begin{pmdkey}
\noindent\textbf{Posterior Identification Theorem.}
Suppose the teacher denoiser is exact and the student's exponential moments $\mathbb{E}_u[e^{\lambda^\top\hat{x}}]$ are finite for all $\lambda\in B_r$. If, for $q_t$-almost every $z_t$,

\begin{equation}
  \frac{\mathbb{E}_u[\hat{x}e^{\lambda^\top\hat{x}}]}
       {\mathbb{E}_u[e^{\lambda^\top\hat{x}}]}
  = D_t(z_t+\rho_t\lambda)
  \label{eq:appB_match}
\end{equation}

holds for every $\lambda\in B_r$, then

\begin{equation}
  p_{G_\theta}(\cdot\mid z_t)
  = q(\cdot\mid z_t)
  \quad\text{for }q_t\text{-a.e. }z_t.
\end{equation}
\end{pmdkey}

\noindent\textit{Proof.}
We use two standard properties of the moment-generating function $\mathbb{E}[e^{\lambda^\top X}]$ of a random vector $X$:

\begin{itemize}
  \item[(i)] On an open set where it is finite, the moment-generating function is infinitely differentiable, with gradient $\mathbb{E}[Xe^{\lambda^\top X}]$. Hence, the gradient of its logarithm is the tilted mean $\mathbb{E}[Xe^{\lambda^\top X}]/\mathbb{E}[e^{\lambda^\top X}]$.

  \item[(ii)] Two random vectors whose moment-generating functions are finite and equal on an open neighborhood of $\lambda=0$ have the same distribution.
\end{itemize}

Fix a noisy state $z_t$ at which the assumptions hold, and define the log moment-generating functions of the student and teacher posterior:

\begin{align*}
  K_{G_\theta}(\lambda)
  &=\log\mathbb{E}_u[e^{\lambda^\top\hat{x}}],\\
  K_q(\lambda)
  &=\log\mathbb{E}_{q(x\mid z_t)}[e^{\lambda^\top x}].
\end{align*}

\noindent\emph{Step 1: Both functions are finite on $B_r$.}
For $K_{G_\theta}$, this follows from the assumption. For $K_q$, Eq.~\eqref{eq:appB_posterior} gives

\begin{equation*}
  \begin{split}
    e^{\lambda^\top x}q(x\mid z_t)
    &\propto q_0(x)\\
    &\quad\times
    \exp\!\left(
      \lambda^\top x
      -\frac{\|z_t-\alpha_t x\|^2}{2\sigma_t^2}
    \right).
  \end{split}
\end{equation*}

The exponential factor is bounded in $x$ because its negative quadratic term dominates the linear term. Hence $\mathbb{E}_{q(x\mid z_t)}[e^{\lambda^\top x}]<\infty$ for every $\lambda\in\mathbb{R}^d$.

\noindent\emph{Step 2: The gradients agree.}
By property (i), $\nabla K_{G_\theta}(\lambda)$ is the left-hand side of Eq.~\eqref{eq:appB_match}, while $\nabla K_q(\lambda)$ is the mean of the tilted posterior $q_\lambda(x\mid z_t)\propto e^{\lambda^\top x}q(x\mid z_t)$. By the Tilt=Shift Identity (Appendix~\ref{app:tilt_shift}) and the exactness of the teacher, this tilted posterior mean equals $D_t(z_t+\rho_t\lambda)$. Therefore, Eq.~\eqref{eq:appB_match} implies

\begin{equation}
  \nabla K_{G_\theta}(\lambda)
  =\nabla K_q(\lambda)
  \quad\text{for all }\lambda\in B_r.
  \label{eq:appB_grad}
\end{equation}

\noindent\emph{Step 3: The moment-generating functions agree.}
Both functions are smooth on $B_r$ by property (i) and vanish at the origin, since $K_{G_\theta}(0)=K_q(0)=\log 1=0$. For any $\lambda\in B_r$, the segment $\{s\lambda:s\in[0,1]\}$ lies in $B_r$. Integrating Eq.~\eqref{eq:appB_grad} along this segment gives

\begin{align*}
  K_{G_\theta}(\lambda)
  &=\int_0^1\lambda^\top
    \nabla K_{G_\theta}(s\lambda)\,ds\\
  &=\int_0^1\lambda^\top
    \nabla K_q(s\lambda)\,ds\\
  &=K_q(\lambda).
\end{align*}

Thus, $\mathbb{E}_u[e^{\lambda^\top\hat{x}}]=\mathbb{E}_{q(x\mid z_t)}[e^{\lambda^\top x}]$ for every $\lambda\in B_r$.

\noindent\emph{Step 4: The distributions agree.}
The moment-generating functions of $p_{G_\theta}(\cdot\mid z_t)$ and $q(\cdot\mid z_t)$ are finite and equal on the open neighborhood $B_r$ of $\lambda=0$. By property (ii), they uniquely determine the same distribution. Therefore,

\begin{equation*}
  p_{G_\theta}(\cdot\mid z_t)
  =q(\cdot\mid z_t).
\end{equation*}

Since the argument applies to every noisy state at which the assumptions hold, posterior equality follows for $q_t$-almost every $z_t$. \hfill$\square$

\section{Interpretations of PMD}
\subsection{PMD as Conditional MMD}
\label{app:mmd}

\noindent\textbf{Setup.}
Fix a noise level $t$ and noisy state $z_t$. Let $q(x\mid z_t)$ denote the teacher posterior and $p_{G_\theta}(x\mid z_t)$ the conditional distribution of the stochastic student $\hat{x}=G_\theta(z_t,u,t)$, where $u\sim\mathcal{N}(0,I)$. For a tilt $\lambda\in\mathbb{R}^d$, let $T_\lambda=D_t(z_t+\rho_t\lambda)$, where $\rho_t=\sigma_t^2/\alpha_t$. Define the feature map

\begin{equation}
  \phi_\lambda(x)
  =
  (x-T_\lambda)
  e^{\lambda^\top(x-T_\lambda)}.
  \label{eq:appC_feature}
\end{equation}

Let $\Pi_{t,z_t}$ denote the tilt distribution, assumed independent of the student parameters, and define the teacher-dependent kernel

\begin{equation}
  k_{t,z_t}(x,y)
  =
  \mathbb{E}_{\lambda\sim\Pi_{t,z_t}}
  \left[
    \phi_\lambda(x)^\top\phi_\lambda(y)
  \right].
  \label{eq:appC_kernel}
\end{equation}

\begin{pmdkey}
\noindent\textbf{Conditional MMD Representation.}
Suppose the teacher denoiser is exact and the required expectations are finite. Then $k_{t,z_t}$ is a positive-semidefinite kernel, and the PMD population objective satisfies

\begin{equation}
  \begin{split}
  \mathcal{L}_{\mathrm{PMD}}(G_\theta)
  &=
  \frac{1}{2}\mathbb{E}_{t,z_t}
  \Big[
    \operatorname{MMD}_{k_{t,z_t}}^2
    \big(
      p_{G_\theta}(\cdot\mid z_t),\\
  &\hspace{35mm}
      q(\cdot\mid z_t)
    \big)
  \Big].
  \end{split}
  \label{eq:appC_mmd}
\end{equation}
\end{pmdkey}

\noindent\textit{Proof.}
We first show that the teacher feature expectation is zero. From the Tilt=Shift Identity (Appendix~\ref{app:tilt_shift}), the teacher target satisfies

\begin{equation}
  T_\lambda
  =
  \frac{
    \mathbb{E}_{q(x\mid z_t)}
    [xe^{\lambda^\top x}]
  }{
    \mathbb{E}_{q(x\mid z_t)}
    [e^{\lambda^\top x}]
  }.
  \label{eq:appC_tilted_mean}
\end{equation}

Multiplying both sides by $\mathbb{E}_{q(x\mid z_t)}[e^{\lambda^\top x}]$ gives

\begin{equation}
  \mathbb{E}_{q(x\mid z_t)}
  [xe^{\lambda^\top x}]
  =
  T_\lambda\,
  \mathbb{E}_{q(x\mid z_t)}
  [e^{\lambda^\top x}].
  \label{eq:appC_tilted_mean_multiplied}
\end{equation}

Rearranging,

\begin{equation}
  \mathbb{E}_{q(x\mid z_t)}
  [xe^{\lambda^\top x}]
  -
  T_\lambda\,
  \mathbb{E}_{q(x\mid z_t)}
  [e^{\lambda^\top x}]
  =0.
  \label{eq:appC_tilted_mean_rearranged}
\end{equation}

Since $T_\lambda$ is constant with respect to $x$, the two terms can be combined into a single expectation:

\begin{equation}
  \mathbb{E}_{q(x\mid z_t)}
  \left[
    (x-T_\lambda)e^{\lambda^\top x}
  \right]
  =0.
  \label{eq:appC_pre_feature_zero}
\end{equation}

Now multiply the integrand by the constant $e^{-\lambda^\top T_\lambda}$. Because this factor does not depend on $x$, Eq.~\eqref{eq:appC_pre_feature_zero} implies

\begin{equation}
  \mathbb{E}_{q(x\mid z_t)}
  \left[
    (x-T_\lambda)
    e^{\lambda^\top x-\lambda^\top T_\lambda}
  \right]
  =0.
  \label{eq:appC_pre_feature_zero_shifted}
\end{equation}

Using
$\lambda^\top x-\lambda^\top T_\lambda
=
\lambda^\top(x-T_\lambda)$,
we obtain

\begin{equation}
  \mathbb{E}_{q(x\mid z_t)}
  \left[
    (x-T_\lambda)
    e^{\lambda^\top(x-T_\lambda)}
  \right]
  =0.
  \label{eq:appC_teacher_zero}
\end{equation}

By the definition of $\phi_\lambda$ in Eq.~\eqref{eq:appC_feature}, this is exactly

\begin{equation}
  \mathbb{E}_{q(x\mid z_t)}
  [\phi_\lambda(x)]
  =0.
  \label{eq:appC_teacher_feature_zero}
\end{equation}

On the student side, writing $h=\hat{x}-T_\lambda$, we have

\begin{equation}
  \mathbb{E}_u[\phi_\lambda(\hat{x})]
  =
  \mathbb{E}_u
  [h e^{\lambda^\top h}]
  =
  r(z_t,\lambda),
  \label{eq:appC_student_residual}
\end{equation}

which is exactly the PMD tilted residual. The kernel in Eq.~\eqref{eq:appC_kernel} is positive semidefinite because it is an expectation of feature inner products. Using its associated feature representation of MMD, we obtain

\begin{equation}
  \begin{split}
  &\operatorname{MMD}_{k_{t,z_t}}^2
  \big(
    p_{G_\theta}(\cdot\mid z_t),
    q(\cdot\mid z_t)
  \big)\\
  &=
  \mathbb{E}_{\lambda\sim\Pi_{t,z_t}}
  \Big[
    \big\|
      \mathbb{E}_u[\phi_\lambda(\hat{x})]
      -
      \mathbb{E}_{q(x\mid z_t)}
      [\phi_\lambda(x)]
    \big\|_2^2
  \Big]\\
  &=
  \mathbb{E}_{\lambda\sim\Pi_{t,z_t}}
  \left[
    \|r(z_t,\lambda)\|_2^2
  \right].
  \end{split}
  \label{eq:appC_mmd_residual}
\end{equation}

Taking the expectation over $(t,z_t)$ and multiplying by $1/2$ recovers the PMD population objective, proving Eq.~\eqref{eq:appC_mmd}. \hfill$\square$

\medskip
\noindent\textbf{Posterior Identification.}
If the student's exponential moments are finite on an open neighborhood of $\lambda=0$ and $\Pi_{t,z_t}$ has positive density there, zero conditional MMD implies $r(z_t,\lambda)=0$ throughout that neighborhood. By the Posterior Identification Theorem (Appendix~\ref{app:identification}), this yields $p_{G_\theta}(\cdot\mid z_t)=q(\cdot\mid z_t)$. Thus, the teacher-dependent kernel is characteristic on this class of conditional distributions. The tilt distribution determines the kernel and therefore the sensitivity of the PMD objective to different posterior discrepancies.

\subsection{PMD as CGF Gradient Matching}
\label{app:cgf}

\noindent\textbf{Setup.}
Fix a noise level $t$ and noisy state $z_t$. Let $q(x\mid z_t)$ denote the teacher posterior and $p_{G_\theta}(x\mid z_t)$ the conditional distribution of the stochastic student $\hat{x}=G_\theta(z_t,u,t)$, where $u\sim\mathcal{N}(0,I)$. Define their cumulant-generating functions (CGFs) as

\begin{equation}
  \begin{split}
  K_{G_\theta}(\lambda)
  &=\log\mathbb{E}_u
    [e^{\lambda^\top\hat{x}}],\\
  K_q(\lambda)
  &=\log\mathbb{E}_{q(x\mid z_t)}
    [e^{\lambda^\top x}].
  \end{split}
  \label{eq:appD_cgf}
\end{equation}

For a tilt $\lambda\in\mathbb{R}^d$, let $T_\lambda=D_t(z_t+\rho_t\lambda)$, where $\rho_t=\sigma_t^2/\alpha_t$. We assume the teacher denoiser is exact and the student's exponential moments are finite on an open neighborhood of $\lambda=0$.

\begin{pmdkey}
\noindent\textbf{CGF Gradient Representation.}
The PMD tilted residual $r(z_t,\lambda)= \mathbb{E}_u[h e^{\lambda^\top h}]$, where $h=\hat{x}-T_\lambda$, satisfies

\begin{equation}
  \begin{split}
  r(z_t,\lambda)
  &=
  e^{K_{G_\theta}(\lambda)-\lambda^\top T_\lambda}\\
  &\quad\times
  \left(
    \nabla_\lambda K_{G_\theta}(\lambda)
    -
    \nabla_\lambda K_q(\lambda)
  \right).
  \end{split}
  \label{eq:appD_residual}
\end{equation}

Consequently, the PMD population objective can be written as

\begin{equation}
  \begin{split}
  \mathcal{L}_{\mathrm{PMD}}(G_\theta)
  &=
  \frac12\mathbb{E}_{t,z_t,\lambda}
  \Big[
    e^{2(K_{G_\theta}(\lambda)
    -\lambda^\top T_\lambda)}\\
  &\quad\times
    \big\|
      \nabla_\lambda K_{G_\theta}(\lambda)
      -
      \nabla_\lambda K_q(\lambda)
    \big\|_2^2
  \Big].
  \end{split}
  \label{eq:appD_loss}
\end{equation}
\end{pmdkey}

\noindent\textit{Proof.}
We first relate the CGF gradients to the exponentially tilted means. Differentiating Eq.~\eqref{eq:appD_cgf} with respect to $\lambda$ gives

\begin{equation}
  \nabla_\lambda K_{G_\theta}(\lambda)
  =
  \frac{
    \mathbb{E}_u
    [\hat{x}e^{\lambda^\top\hat{x}}]
  }{
    \mathbb{E}_u
    [e^{\lambda^\top\hat{x}}]
  },
  \label{eq:appD_student_grad}
\end{equation}

and similarly,

\begin{equation}
  \nabla_\lambda K_q(\lambda)
  =
  \frac{
    \mathbb{E}_{q(x\mid z_t)}
    [xe^{\lambda^\top x}]
  }{
    \mathbb{E}_{q(x\mid z_t)}
    [e^{\lambda^\top x}]
  }.
  \label{eq:appD_teacher_grad}
\end{equation}

By the Tilt=Shift Identity (Appendix~\ref{app:tilt_shift}), the teacher's tilted mean equals $T_\lambda$. Hence,

\begin{equation}
  \nabla_\lambda K_q(\lambda)
  =
  T_\lambda
  =
  D_t(z_t+\rho_t\lambda).
  \label{eq:appD_teacher_identity}
\end{equation}

Now consider the PMD residual. Substituting $h=\hat{x}-T_\lambda$ gives

\begin{equation}
  \begin{split}
  r(z_t,\lambda)
  &=
  \mathbb{E}_u
  \left[
    (\hat{x}-T_\lambda)
    e^{\lambda^\top(\hat{x}-T_\lambda)}
  \right]\\
  &=
  e^{-\lambda^\top T_\lambda}
  \mathbb{E}_u
  \left[
    (\hat{x}-T_\lambda)
    e^{\lambda^\top\hat{x}}
  \right].
  \end{split}
  \label{eq:appD_residual_expand}
\end{equation}

Since $T_\lambda$ is independent of $u$, we can separate the expectation:

\begin{equation}
  \begin{split}
  r(z_t,\lambda)
  &=
  e^{-\lambda^\top T_\lambda}
  \Big(
    \mathbb{E}_u
    [\hat{x}e^{\lambda^\top\hat{x}}]\\
  &\qquad-
    T_\lambda
    \mathbb{E}_u
    [e^{\lambda^\top\hat{x}}]
  \Big).
  \end{split}
  \label{eq:appD_residual_separate}
\end{equation}

From Eq.~\eqref{eq:appD_student_grad},

\begin{equation}
  \begin{split}
  &\mathbb{E}_u
  [\hat{x}e^{\lambda^\top\hat{x}}]\\
  &\quad=
  \mathbb{E}_u
  [e^{\lambda^\top\hat{x}}]\,
  \nabla_\lambda K_{G_\theta}(\lambda).
  \end{split}
  \label{eq:appD_student_moment}
\end{equation}

Substituting into Eq.~\eqref{eq:appD_residual_separate} and factoring out the common term yields

\begin{equation}
  \begin{split}
  r(z_t,\lambda)
  &=
  e^{-\lambda^\top T_\lambda}
  \mathbb{E}_u[e^{\lambda^\top\hat{x}}]\\
  &\quad\times
  \left(
    \nabla_\lambda K_{G_\theta}(\lambda)
    -
    T_\lambda
  \right).
  \end{split}
  \label{eq:appD_residual_factor}
\end{equation}

By the definition of the student CGF,

\begin{equation}
  \mathbb{E}_u[e^{\lambda^\top\hat{x}}]
  =
  e^{K_{G_\theta}(\lambda)}.
  \label{eq:appD_mgf}
\end{equation}

Combining this with Eq.~\eqref{eq:appD_teacher_identity} gives

\begin{equation}
  \begin{split}
  r(z_t,\lambda)
  &=
  e^{K_{G_\theta}(\lambda)-\lambda^\top T_\lambda}\\
  &\quad\times
  \left(
    \nabla_\lambda K_{G_\theta}(\lambda)
    -
    \nabla_\lambda K_q(\lambda)
  \right),
  \end{split}
  \label{eq:appD_residual_final}
\end{equation}

which proves Eq.~\eqref{eq:appD_residual}. Substituting this expression into the PMD population objective

\begin{equation*}
  \mathcal{L}_{\mathrm{PMD}}(G_\theta)
  =
  \frac12\mathbb{E}_{t,z_t,\lambda}
  [\|r(z_t,\lambda)\|_2^2]
\end{equation*}

immediately gives Eq.~\eqref{eq:appD_loss}.
\hfill$\square$

\medskip
\noindent\textbf{The Teacher CGF from Denoiser Evaluations.} Equation~\eqref{eq:appD_teacher_identity} provides an additional interpretation of the Tilt=Shift Identity. Since $K_q(0)=0$, integrating the teacher CGF gradient along the line from $0$ to $\lambda$ gives

\begin{equation}
  \begin{split}
  K_q(\lambda)
  &=
  \int_0^1
  \lambda^\top\nabla_\lambda K_q(s\lambda)
  \,ds\\
  &=
  \int_0^1
  \lambda^\top
  D_t(z_t+s\rho_t\lambda)
  \,ds.
  \end{split}
  \label{eq:appD_teacher_cgf}
\end{equation}

Thus, an exact denoising teacher implicitly specifies the entire posterior CGF through its predictions at shifted noisy inputs. In particular, its derivatives at $\lambda=0$ recover the posterior mean and covariance:

\begin{equation}
  \nabla_\lambda K_q(0)=D_t(z_t),
  \label{eq:appD_mean}
\end{equation}

\begin{equation}
  \begin{split}
  \nabla_\lambda^2 K_q(0)
  &=
  \rho_t\nabla_{z_t}D_t(z_t)\\
  &=
  \operatorname{Cov}_{q(x\mid z_t)}(x).
  \end{split}
  \label{eq:appD_covariance}
\end{equation}

Higher derivatives similarly recover higher-order posterior cumulants. PMD therefore accesses information about the full posterior through shifted denoiser evaluations, without explicitly estimating these higher-order statistics.

\medskip
\noindent\textbf{Posterior Identification.}
The exponential factor in Eq.~\eqref{eq:appD_residual} is strictly positive wherever the CGFs are finite. Consequently,

\begin{equation}
  r(z_t,\lambda)=0
  \quad\Longleftrightarrow\quad
  \nabla_\lambda K_{G_\theta}(\lambda)
  =
  \nabla_\lambda K_q(\lambda).
  \label{eq:appD_zero}
\end{equation}

If this equality holds throughout an open neighborhood of $\lambda=0$, the two CGFs differ only by a constant. Since $K_{G_\theta}(0)=K_q(0)=0$, they coincide on that neighborhood:

\begin{equation}
  K_{G_\theta}(\lambda)=K_q(\lambda).
  \label{eq:appD_cgf_equal}
\end{equation}

By uniqueness of moment-generating functions, the student and teacher conditional distributions are identical, as established in Appendix~\ref{app:identification}.

Thus, PMD matches the gradients of the student and teacher CGFs through a weighted squared discrepancy. Unlike matching a finite collection of moments, matching these gradients over an open neighborhood identifies the full conditional posterior. The weighting factor depends on the student distribution, but does not change the zero set of the matching conditions.

\subsection{PMD as MGF-Ratio Gradient Matching}
\label{app:mgf_ratio}

\noindent\textbf{Setup.}
Fix a noise level $t$ and noisy state $z_t$. Let $q(x\mid z_t)$ denote the teacher posterior and $p_{G_\theta}(x\mid z_t)$ the conditional distribution of the stochastic student $\hat{x}=G_\theta(z_t,u,t)$, where $u\sim\mathcal{N}(0,I)$. Define their moment-generating functions (MGFs) as

\begin{equation}
  \begin{split}
  M_{G_\theta}(\lambda)
  &=
  \mathbb{E}_u[e^{\lambda^\top\hat{x}}],\\
  M_q(\lambda)
  &=
  \mathbb{E}_{q(x\mid z_t)}
  [e^{\lambda^\top x}].
  \end{split}
  \label{eq:appE_mgfs}
\end{equation}

For a tilt $\lambda\in\mathbb{R}^d$, let $T_\lambda=D_t(z_t+\rho_t\lambda)$, where $\rho_t=\sigma_t^2/\alpha_t$. We assume the teacher denoiser is exact and the student's exponential moments are finite on an open neighborhood of $\lambda=0$. All expressions below are considered where the required expectations are finite.

\begin{pmdkey}
\noindent\textbf{MGF-Ratio Gradient Representation.}
The PMD tilted residual
$r(z_t,\lambda)
=\mathbb{E}_u[h e^{\lambda^\top h}]$,
where $h=\hat{x}-T_\lambda$, satisfies

\begin{equation}
  \begin{split}
  r(z_t,\lambda)
  &=
  w_q(\lambda)
  \nabla_\lambda
  \left(
    \frac{M_{G_\theta}(\lambda)}
         {M_q(\lambda)}
  \right),\\
  w_q(\lambda)
  &=
  e^{-\lambda^\top T_\lambda}
  M_q(\lambda)>0.
  \end{split}
  \label{eq:appE_residual}
\end{equation}

Consequently, the PMD population objective admits the exact representation

\begin{equation}
  \begin{split}
  \mathcal{L}_{\mathrm{PMD}}(G_\theta)
  &=
  \frac12\mathbb{E}_{t,z_t,\lambda}
  \Bigg[
    w_q(\lambda)^2\\
  &\quad\times
    \left\|
      \nabla_\lambda
      \left(
        \frac{M_{G_\theta}(\lambda)}
             {M_q(\lambda)}
      \right)
    \right\|_2^2
  \Bigg].
  \end{split}
  \label{eq:appE_loss}
\end{equation}
\end{pmdkey}

\noindent\textit{Proof.}
We first express the teacher target in terms of the teacher MGF. By the Tilt=Shift Identity (Appendix~\ref{app:tilt_shift}),

\begin{equation}
  T_\lambda
  =
  \frac{
    \mathbb{E}_{q(x\mid z_t)}
    [xe^{\lambda^\top x}]
  }{
    \mathbb{E}_{q(x\mid z_t)}
    [e^{\lambda^\top x}]
  }.
  \label{eq:appE_tilted_mean}
\end{equation}

Differentiating $M_q(\lambda)$ with respect to $\lambda$ gives

\begin{equation}
  \nabla_\lambda M_q(\lambda)
  =
  \mathbb{E}_{q(x\mid z_t)}
  [xe^{\lambda^\top x}].
  \label{eq:appE_teacher_derivative}
\end{equation}

Therefore,

\begin{equation}
  T_\lambda
  =
  \frac{\nabla_\lambda M_q(\lambda)}
       {M_q(\lambda)}.
  \label{eq:appE_teacher_ratio}
\end{equation}

Now consider the PMD residual. Substituting $h=\hat{x}-T_\lambda$,

\begin{equation}
  \begin{split}
  r(z_t,\lambda)
  &=
  \mathbb{E}_u
  \left[
    (\hat{x}-T_\lambda)
    e^{\lambda^\top(\hat{x}-T_\lambda)}
  \right]\\
  &=
  e^{-\lambda^\top T_\lambda}
  \mathbb{E}_u
  \left[
    (\hat{x}-T_\lambda)
    e^{\lambda^\top\hat{x}}
  \right].
  \end{split}
  \label{eq:appE_residual_expand}
\end{equation}

Since $T_\lambda$ is independent of $u$, we can separate the expectation:

\begin{equation}
  \begin{split}
  r(z_t,\lambda)
  &=
  e^{-\lambda^\top T_\lambda}
  \Big(
    \mathbb{E}_u
    [\hat{x}e^{\lambda^\top\hat{x}}]\\
  &\qquad-
    T_\lambda
    \mathbb{E}_u
    [e^{\lambda^\top\hat{x}}]
  \Big).
  \end{split}
  \label{eq:appE_residual_separate}
\end{equation}

By the definition of the student MGF,

\begin{equation}
  \begin{split}
  \nabla_\lambda M_{G_\theta}(\lambda)
  &=
  \mathbb{E}_u
  [\hat{x}e^{\lambda^\top\hat{x}}].
  \end{split}
  \label{eq:appE_student_derivative}
\end{equation}

Substituting Eqs.~\eqref{eq:appE_teacher_ratio} and~\eqref{eq:appE_student_derivative} into Eq.~\eqref{eq:appE_residual_separate} yields

\begin{equation}
  \begin{split}
  r(z_t,\lambda)
  &=
  e^{-\lambda^\top T_\lambda}
  \Bigg(
    \nabla_\lambda M_{G_\theta}(\lambda)\\
  &\quad-
    \frac{M_{G_\theta}(\lambda)}
         {M_q(\lambda)}
    \nabla_\lambda M_q(\lambda)
  \Bigg).
  \end{split}
  \label{eq:appE_residual_mgfs}
\end{equation}

Applying the quotient rule,

\begin{equation}
  \begin{split}
  &\nabla_\lambda
  \left(
    \frac{M_{G_\theta}(\lambda)}
         {M_q(\lambda)}
  \right)\\
  &=
  \frac{
    \nabla_\lambda M_{G_\theta}(\lambda)
  }{
    M_q(\lambda)
  }\\
  &\quad-
  \frac{
    M_{G_\theta}(\lambda)
  }{
    M_q(\lambda)^2
  }
  \nabla_\lambda M_q(\lambda).
  \end{split}
  \label{eq:appE_quotient}
\end{equation}

Multiplying Eq.~\eqref{eq:appE_quotient} by $M_q(\lambda)$, the expression in parentheses in Eq.~\eqref{eq:appE_residual_mgfs} becomes

\begin{equation}
  \begin{split}
  &\nabla_\lambda M_{G_\theta}(\lambda)
  -
  \frac{M_{G_\theta}(\lambda)}
       {M_q(\lambda)}
  \nabla_\lambda M_q(\lambda)\\
  &=
  M_q(\lambda)
  \nabla_\lambda
  \left(
    \frac{M_{G_\theta}(\lambda)}
         {M_q(\lambda)}
  \right).
  \end{split}
  \label{eq:appE_quotient_result}
\end{equation}

Substituting back, we obtain

\begin{equation}
  \begin{split}
  r(z_t,\lambda)
  &=
  e^{-\lambda^\top T_\lambda}
  M_q(\lambda)\\
  &\quad\times
  \nabla_\lambda
  \left(
    \frac{M_{G_\theta}(\lambda)}
         {M_q(\lambda)}
  \right).
  \end{split}
  \label{eq:appE_residual_final}
\end{equation}

Defining $w_q(\lambda)= e^{-\lambda^\top T_\lambda}M_q(\lambda)$ proves Eq.~\eqref{eq:appE_residual}. Substituting this identity into

\begin{equation*}
  \mathcal{L}_{\mathrm{PMD}}(G_\theta)
  =
  \frac12\mathbb{E}_{t,z_t,\lambda}
  [\|r(z_t,\lambda)\|_2^2]
\end{equation*}

gives Eq.~\eqref{eq:appE_loss}.
\hfill$\square$

\medskip
\noindent\textbf{Teacher-Dependent Weighting.}
Unlike the CGF-gradient representation in
Appendix~\ref{app:cgf}, the weighting factor
$w_q(\lambda)$ depends only on the teacher
posterior. Writing
$K_q(\lambda)=\log M_q(\lambda)$,

\begin{equation}
  w_q(\lambda)
  =
  e^{K_q(\lambda)-\lambda^\top T_\lambda}.
  \label{eq:appE_weight}
\end{equation}

Since $K_q$ is convex, $K_q(0)=0$, and $\nabla_\lambda K_q(\lambda)=T_\lambda$, the supporting hyperplane inequality gives $K_q(\lambda)\leq\lambda^\top T_\lambda$. Therefore,

\begin{equation}
  0<w_q(\lambda)\leq 1.
  \label{eq:appE_weight_bound}
\end{equation}

Thus, PMD can be interpreted as minimizing a teacher-weighted squared gradient of the ratio between the student and teacher MGFs.

\medskip
\noindent\textbf{Posterior Identification.}
Suppose the student's exponential moments
are finite on an open ball $B_r$ around zero
and the tilt distribution has positive density
throughout $B_r$.
If the conditional PMD loss is zero, then
Eq.~\eqref{eq:appE_loss}, together with
$w_q(\lambda)>0$, implies

\begin{equation}
  \nabla_\lambda
  \left(
    \frac{M_{G_\theta}(\lambda)}
         {M_q(\lambda)}
  \right)
  =0
  \quad \text{for all }\lambda\in B_r.
  \label{eq:appE_zero_gradient}
\end{equation}

Here, equality initially holds almost everywhere under the tilt distribution and extends throughout $B_r$ by continuity. Since $B_r$ is connected, the MGF ratio must be constant on this neighborhood.

Both MGFs equal one at $\lambda=0$:

\begin{equation}
  M_{G_\theta}(0)=M_q(0)=1.
  \label{eq:appE_normalization}
\end{equation}

Hence their ratio is identically one, giving

\begin{equation}
  M_{G_\theta}(\lambda)=M_q(\lambda)
  \quad\text{for all }\lambda\in B_r.
  \label{eq:appE_mgf_equal}
\end{equation}

By uniqueness of moment-generating functions,

\begin{equation}
  p_{G_\theta}(\cdot\mid z_t)
  =
  q(\cdot\mid z_t),
  \label{eq:appE_posterior_equal}
\end{equation}

recovering the Posterior Identification Theorem (Appendix~\ref{app:identification}).

This representation therefore provides an alternative interpretation of PMD: the loss penalizes variations in the student-to-teacher MGF ratio, and its zero-loss condition forces that ratio to equal one. The ratio and its weighting factor are used only for theoretical interpretation; PMD training evaluates the original tilted residual directly, without explicitly computing either MGF.

\section{Implementation Details}
\label{sec:appendix_implementation_details}
\subsection{Shared Training Procedure}
\label{app:shared_training}

\paragraph{Initialization and stochastic inputs.}
For both backbones, we initialize the student and frozen teacher from the corresponding undistilled base model. The VAE and text encoders remain frozen. At each student call, we mix the current latent with independent Gaussian noise and concatenate this mixed latent with the original state. We widen only the first input layer, retaining its pretrained weights and initializing the added raw-state weights to zero. The output shape is unchanged. Independent noise draws make the student stochastic; the added input channels preserve access to the original conditioning state.

\paragraph{On-policy training and adversarial loss.}
We obtain detached training states by rolling out the current student. At each selected noise level, the same four independent student draws are reused across four tilted teacher queries, using spatially coarse probes. We combine PMD with a non-saturating logistic GAN loss of fixed weight $0.75$. A trainable discriminator head reads caption-conditioned features from the frozen teacher. The discriminator uses independently noised real and generated latents at integer levels sampled uniformly from $\{1,\ldots,999\}$, under each backbone's native corruption schedule. Caption and reference-image streams are sampled independently. We perform five discriminator updates on fresh batches before each student update, using global batch size 16 without student EMA. To limit memory, we cache student outputs, compute their loss gradients, and replay each draw with the same noise for backpropagation. Tables~\ref{tab:sdxl_training_hyperparameters} and~\ref{tab:flux_training_hyperparameters} give the remaining settings.

\paragraph{Finite-sample bias correction.}
A softmax-weighted mean is a self-normalized ratio estimate and is generally biased with finitely many draws. Squaring a single noisy residual also adds a variance term. Algorithm~\ref{alg:pmd_loss_finite_sample} extends Algorithm~\ref{alg:pmd_loss} for $M\geq4$ independent draws, replacing its weighted-mean and squared-error steps. For a subset $S$ of $n\geq2$ draws, let $h_i=\hat{x}^{(i)}-T_\lambda$ and define
\begin{equation}
  \begin{aligned}
    r(S)&=\sum_{i\in S}
      \frac{\exp(\lambda^\top\hat{x}^{(i)})}
           {\sum_{j\in S}\exp(\lambda^\top\hat{x}^{(j)})}\,h_i,\\
    r_{\mathrm{JK}}(S)&=n\,r(S)
      -\frac{n-1}{n}\sum_{i\in S}r(S\setminus\{i\}).
  \end{aligned}
  \label{eq:app_group_jackknife}
\end{equation}
We cross-fit two disjoint groups and average over the set $\mathcal P$ of all unique balanced partitions of the $M$ draws:
\begin{equation}
  \widehat{\mathcal L}_{\mathrm{PMD}}
  =\frac{1}{|\mathcal P|}\sum_{(A,B)\in\mathcal P}
    r_{\mathrm{JK}}(A)^\top r_{\mathrm{JK}}(B).
  \label{eq:app_crossfit_pmd}
\end{equation}
The groups have sizes $\lfloor M/2\rfloor$ and $\lceil M/2\rceil$; for our default $M=4$, there are three partitions into pairs. The delete-one jackknife reduces the leading ratio bias, while cross-fitting avoids the extra variance term from squaring a noisy residual. Both factors and their softmax weights remain differentiable. The estimator is not generally unbiased at nonzero tilt; at zero tilt it is the unbiased squared-mean U-statistic. Signed estimates are not clamped.

\begin{algorithm}[H]
\caption{Finite-sample correction to Algorithm~\ref{alg:pmd_loss}}
\label{alg:pmd_loss_finite_sample}
\begin{lstlisting}[style=pmd]
# Reuse x_hat, tilt, target from Algorithm 1
M = len(x_hat)  # M >= 4 independent draws
h = [x - target for x in x_hat]
logits = [dot(tilt, x) for x in x_hat]

def residual(S):
    w = softmax([logits[i] for i in S])
    return sum(w[k] * h[i]
               for k, i in enumerate(S))

def corrected_residual(S):
    n = len(S)
    leave_one_out = mean([residual(S - {i})
                          for i in S])
    return n * residual(S) - (n-1) * leave_one_out

# All unique balanced partitions of the M draws
splits = balanced_splits(M)
return mean([dot(corrected_residual(A),
                 corrected_residual(B))
             for A, B in splits])
\end{lstlisting}
\end{algorithm}

\subsection{SDXL}
\label{app:sdxl_training}

\begin{table*}[t]
\centering
\small
\setlength{\tabcolsep}{6pt}
\renewcommand{\arraystretch}{1.08}
\begin{tabular}{@{}p{0.30\textwidth}p{0.66\textwidth}@{}}
\toprule
\textbf{Setting} & \textbf{Value} \\
\midrule
Student initialization / frozen teacher & SDXL Base 1.0 / same base model \\
Image resolution & $1024\times1024$ \\
Step Schedule & $\{999,\,749,\,499,\,249\}$ \\
Stochastic student noise dials & $\{0.90,\,0.30,\,0.45,\,0.15\}$ at the four sampling levels \\
Tilt probes & Coarse \\
Normalized tilt radius & Uniform on $[0.25,0.50]$ \\
Student draws / PMD target queries & $4\,/\,4$ \\
Global caption batch size & 16 \\
Learning rates (student / discriminator) & $5\times10^{-7}\,/\,5\times10^{-5}$ \\
Optimizer (both) & AdamW; $\beta=(0.9,0.999)$, $\epsilon=10^{-8}$, weight decay 0 \\
Gradient norm clipping & 10 for both student and discriminator \\
Discriminator updates per student update & 5 \\
Auxiliary GAN loss & Non-saturating logistic loss; fixed student weight $0.75$ \\
Teacher / student guidance scale & $1\,/\,1$ \\
Backward simulation & Yes; rollout states are detached \\
Student EMA & None \\
Training length & 2,000 student updates and 10,000 discriminator updates \\
\bottomrule
\end{tabular}
\caption{\textbf{SDXL training hyperparameters.} Step Schedule uses the native discrete VP noise labels. Training draws are separate from the eight full generations per prompt used for diversity evaluation.}
\label{tab:sdxl_training_hyperparameters}
\end{table*}

\paragraph{Backbone and clean estimates.}
We use SDXL Base 1.0~\cite{podell2024sdxl} at $1024\times1024$ resolution, with latent shape $4\times128\times128$. Concatenating the mixed and original latents expands the U-Net input convolution from 4 to 8 channels; its output remains 4 channels. SDXL predicts noise $\epsilon_\theta$, which we convert to a clean latent using its native variance-preserving (VP) schedule. For a mixed input $y$ evaluated at level $\tilde t$,
\begin{equation}
  \hat x=\frac{y-\sigma_{\tilde t}\,
    \epsilon_\theta([y,z_t],\tilde t,c)}{\alpha_{\tilde t}},
  \label{eq:app_sdxl_clean}
\end{equation}
where $c$ is the caption conditioning and $\alpha_t^2+\sigma_t^2=1$. At noninitial steps, the effective level $\tilde t$ accounts for the additional mixed-in noise; the first call retains label 999 and its native coefficients. Frozen teacher targets use the same noise-to-clean conversion at the queried level.

\paragraph{Data, tilts, and sampling.}
We draw both student training prompts and GAN reference image--caption pairs from LAION, caching scaled VAE latents and dual-CLIP text embeddings. The discriminator head reads the frozen SDXL encoder and middle-block features. The four coarse tilt directions and their radii are sampled independently, with probes formed on a $16\times16$ grid. We supervise all four levels in Table~\ref{tab:sdxl_training_hyperparameters}. Between sampling steps, we re-noise the predicted clean latent using the next level's VP coefficients and fresh Gaussian noise.

\subsection{FLUX.2 [klein] 4B}
\label{app:flux_training}

\begin{table*}[t]
\centering
\small
\setlength{\tabcolsep}{6pt}
\renewcommand{\arraystretch}{1.08}
\begin{tabular}{@{}p{0.30\textwidth}p{0.66\textwidth}@{}}
\toprule
\textbf{Setting} & \textbf{Value} \\
\midrule
Student initialization / frozen teacher & Undistilled FLUX.2 [klein] Base 4B / same base model \\
Image resolution & $1024\times1024$ \\
Step Schedule & $\{1000,\,958.09,\,883.98,\,717.50\}$ \\
Stochastic student noise dials & $\{0,\,0.30,\,0.45,\,0.15\}$ at the four sampling levels \\
Tilt probes & Coarse \\
Normalized tilt radius & Uniform on $[0.25,0.50]$ \\
Student draws / PMD target queries & $4\,/\,4$ \\
Global caption batch size & 16 \\
Learning rates (student / discriminator) & $5\times10^{-7}\,/\,5\times10^{-5}$ \\
Optimizer (both) & AdamW; $\beta=(0.9,0.999)$, $\epsilon=10^{-8}$, weight decay 0 \\
Gradient norm clipping & 10 for both student and discriminator \\
Discriminator updates per student update & 5 \\
Auxiliary GAN loss & Non-saturating logistic loss; fixed student weight $0.75$ \\
Teacher / student guidance scale & $1\,/\,1$ \\
Backward simulation & Yes; rollout states are detached \\
Student EMA & None \\
Training length & 2,000 student updates and 10,000 discriminator updates \\
\bottomrule
\end{tabular}
\caption{\textbf{FLUX.2 [klein] 4B training hyperparameters.} The native $1024\times1024$ step schedule is reported as $1000t$. Training draws are separate from the eight full generations per prompt used for diversity evaluation.}
\label{tab:flux_training_hyperparameters}
\end{table*}

\paragraph{Backbone and clean estimates.}
We use the undistilled FLUX.2 [klein] Base 4B model~\cite{flux-2-2025}. Concatenating the tokenized mixed and original latents expands the input projection from 128 to 256 features per image token; the hidden size (3072) and output size (128) remain unchanged. We express the noise mixing in rectified-flow (RF) coordinates. Unlike SDXL's noise prediction, FLUX predicts velocity $v_\theta$, giving the clean estimate
\begin{equation}
  \hat x=y-\tilde t\,v_\theta([y,z_t],\tilde t,c),
  \label{eq:app_flux_clean}
\end{equation}
where $y$ and $\tilde t$ are the mixed latent and its effective RF noise level. The frozen teacher uses the same velocity-to-clean conversion.

\paragraph{Data, tilts, and sampling.}
Training uses FLUX-Reason-6M~\cite{fang2026flux}, excluding its validation split. The discriminator head reads the frozen teacher's midpoint image-token features. Tilt queries use coarse probes formed on an $8\times8$ grid. We use the native $1024\times1024$ sampling schedule, with internal RF levels $t\in[0,1]$; Table~\ref{tab:flux_training_hyperparameters} reports $1000t$, rounded to two decimal places. The initial pure-noise call is used for rollout and inference without a direct PMD loss; training samples one of the three remaining levels per update. Inference uses the four-step CPS sampler with $\eta=0.9$ and guidance scale 1.

\medskip
\begin{samepage}
\subsection{Details on Evaluation Metrics}
\label{app:eval_metrics}

\paragraph{Diversity protocol.}
For each caption, we generate $K=8$ images in eight complete sampling runs with different random seeds. The main comparison and tilt ablations uses all 1,632 PartiPrompts~\cite{yu2022scaling}. For a pairwise distance $d$, we average all $\binom{8}{2}=28$ unordered pairs within each prompt, then average over the $P$ prompts:
\begin{equation}
  \mathcal{D}_d
  =\frac{1}{P}\sum_{p=1}^{P}
    \frac{2}{K(K-1)}\sum_{i<j}d(x_{p,i},x_{p,j}).
  \label{eq:app_metric_pairwise}
\end{equation}
All methods use matched prompts and seeds. Higher diversity scores indicate greater sample variation, and should be considered alongside quality and prompt adherence.
\end{samepage}

\paragraph{Vendi score.}
Vendi~\cite{friedman2022vendi} summarizes the effective number of different samples in a set. For each prompt, we form the cosine-similarity matrix $S_{ij}=f_i^\top f_j$ from unit-normalized DINOv2 features. If $s_1,\ldots,s_K$ are the eigenvalues of $S/K$, then
\begin{equation}
  \operatorname{Vendi}
  =\exp\!\left(-\sum_{i=1}^{K}s_i\log s_i\right),
  \qquad 0\log 0:=0.
  \label{eq:app_metric_vendi}
\end{equation}
We average this set-level score over prompts. Its range is $[1,8]$: identical features give 1, while orthogonal features give 8.

\paragraph{DINOv2 diversity.}
We extract the CLS embedding from DINOv2 ViT-B/14~\cite{oquab2023dinov2} using its official image preprocessing. For an encoder $f$, the cosine distance is
\begin{equation}
  d_f(x,y)
  =1-\frac{f(x)^\top f(y)}{\|f(x)\|_2\|f(y)\|_2}.
  \label{eq:app_metric_cosine}
\end{equation}
Applying Eq.~\eqref{eq:app_metric_pairwise} with these features measures variation in image-level visual representations.

\paragraph{DreamSim diversity.}
DreamSim~\cite{fu2023dreamsim} is trained to reflect human judgments of visual similarity, including appearance, pose, and layout. We use the released ensemble of DINO, CLIP, and OpenCLIP with its official $224\times224$ preprocessing. We compute cosine distances between its normalized ensemble embeddings using Eq.~\eqref{eq:app_metric_cosine}, then average all within-prompt pairs as in Eq.~\eqref{eq:app_metric_pairwise}.

\begin{table*}[t]
    \centering
    \begin{tabular*}{\textwidth}{@{\extracolsep{\fill}}lccccc@{}}
        \toprule
        \textbf{Tilt Radius} & \textbf{Vendi} $\uparrow$ & \textbf{DreamSim} $\uparrow$ & \textbf{DINOv2} $\uparrow$ & \textbf{LPIPS} $\uparrow$ & \textbf{FID COCO-10k} $\downarrow$ \\
        \midrule
        $\mathcal{U}(0.25, 0.5)$ & 3.370 & 0.3503 & 0.3690 & 0.5705 & \cellcolor{orange!10}17.47 \\
        $\mathcal{U}(0.5, 1)$ & 3.393 & \cellcolor{orange!10}0.3755 & 0.3742 & \cellcolor{orange!10}0.5928 & 17.76 \\
        $\mathcal{U}(1, 2)$ & \cellcolor{orange!20}\textbf{4.269} & \cellcolor{orange!20}\textbf{0.4353} & \cellcolor{orange!20}\textbf{0.4880} & \cellcolor{orange!20}\textbf{0.6220} & 17.87 \\
        $\mathcal{U}(0.25, 2)$ & \cellcolor{orange!10}3.559 & 0.3723 & \cellcolor{orange!10}0.3982 & 0.5755 & \cellcolor{orange!20}\textbf{16.40} \\
        \bottomrule
    \end{tabular*}
    \caption{\textbf{Ablation study on the tilt radius.} Tilt radii are uniformly sampled from the specified intervals. Higher Vendi, DreamSim, DINOv2, and LPIPS scores indicate greater sample diversity, while lower FID indicates better generation quality. Darker and lighter orange indicate the best and second-best results, respectively.}
    \label{tab:tilt_radius_ablation}
\end{table*}

\paragraph{LPIPS diversity.}
LPIPS~\cite{zhang2018unreasonable} compares spatial feature maps using learned channel weights. Here, $x$ and $y$ are two generated images for the same caption:
\begin{equation}
  d_{\mathrm{LPIPS}}(x,y)
  =\sum_{\ell,c}a_{\ell c}
    \frac{\|\hat f_{\ell c}(x)-\hat f_{\ell c}(y)\|_F^2}
         {H_\ell W_\ell},
  \label{eq:app_metric_lpips}
\end{equation}
where $\hat f$ denotes channel-normalized features, $a$ the learned weights, and $H_\ell W_\ell$ the spatial size at layer $\ell$. We use AlexNet LPIPS v0.1, resize images to $256\times256$ with bicubic interpolation, scale RGB values to $[-1,1]$, and average all 28 generated-image pairs per caption using Eq.~\eqref{eq:app_metric_pairwise}. Larger LPIPS therefore indicates greater intra-prompt diversity. In reconstruction evaluation, the distance to a reference is instead minimized. We use LPIPS only for evaluation, not as a training loss.

\paragraph{Prompt-following benchmarks.}
We use each benchmark's own prompts and four images per prompt. GenEval~\cite{ghosh2023geneval} uses 553 prompts and object detection to check object presence, count, color, and spatial relations; we report the macro-average over its six tasks. DPG-Bench~\cite{hu2024ella} uses 1,065 prompts and mPLUG to answer yes/no questions about each image. Answers to dependent questions receive zero credit when a prerequisite fails; scores are averaged over questions, images, and prompts. OneIG-EN~\cite{chang2026oneig} uses 1,120 English prompts and its released evaluators; we report the arithmetic mean of alignment, text rendering, reasoning, style, and diversity scores.

\paragraph{Image--text alignment and preference.}
For these metrics, we use a fixed set of 10,000 COCO 2014 validation captions, with one generated image per caption and matched seeds across models.  HPSv3~\cite{ma2025hpsv3} is learned human-preference scorer; we average its released reward outputs over image--caption pairs. For HPSv3, we use the predicted mean reward. PickScore~\cite{kirstain2023pick} is the learned temperature-scaled image--text cosine score, without a candidate-set softmax. CLIP-S~\cite{radford2021learning} is $100\cos(f_I(x),f_T(p))$, using OpenCLIP ViT-g/14, the text prefix ``A photo depicts'', and a $256$-pixel image pre-resize. Higher scores are better for all four metrics.

\begin{figure*}[t]
    \centering
    \includegraphics[width=1.0\linewidth]{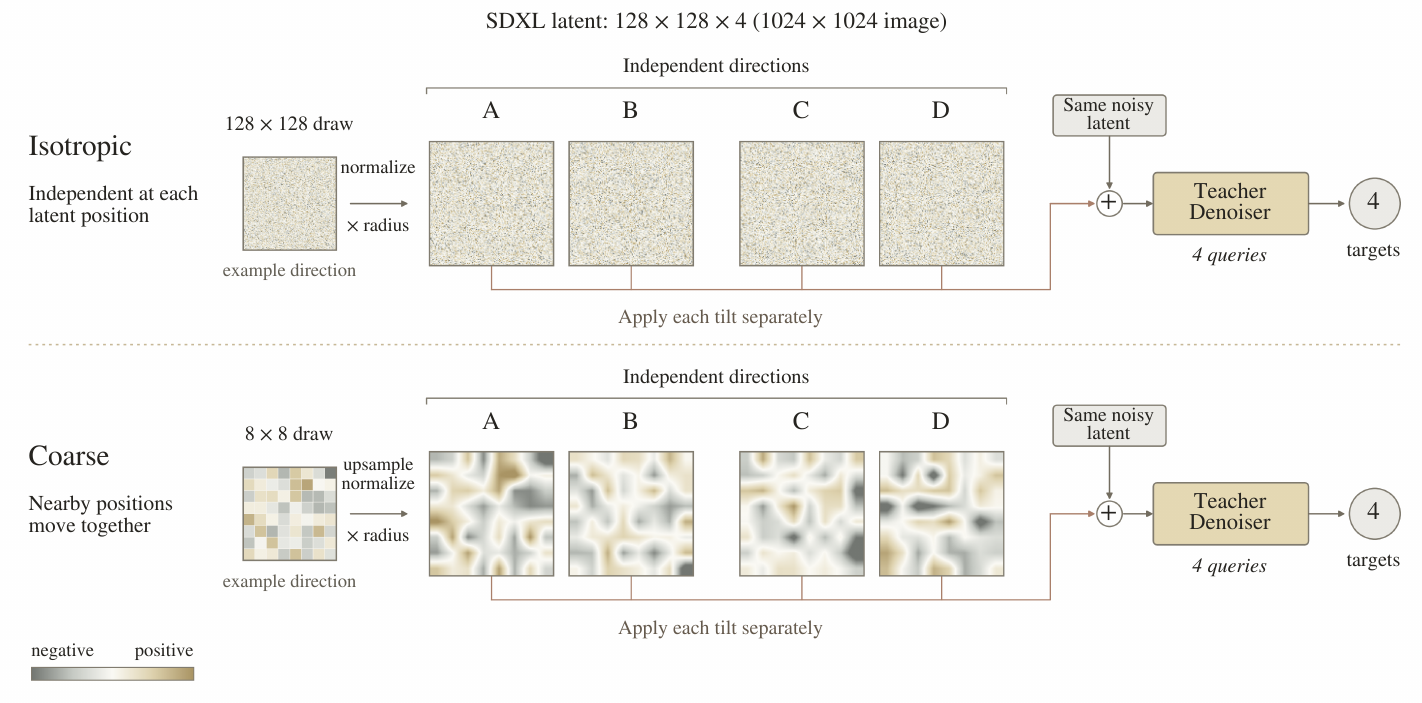} \\
    \caption{\textbf{Tilt selection ablation.} Comparison of isotropic and coarse spatially correlated tilt directions in SDXL latent space. Four independent tilts are applied to the same noisy latent, each producing a separate teacher target.}
    \label{app:tilt_selection_ablation}
\end{figure*}

\begin{table*}[t]
\centering
\small
\setlength{\tabcolsep}{4pt}
\renewcommand{\arraystretch}{1.12}
\begin{tabular}{@{}l*{7}{c}@{}}
\toprule
\textbf{Training objective} & ImageReward $\uparrow$ & CLIP-S $\uparrow$ & FID $\downarrow$ & Vendi $\uparrow$ & DreamSim $\uparrow$ & DINOv2 $\uparrow$ & LPIPS $\uparrow$ \\
\midrule
PMD only
& -1.232 & 26.88 & 56.13
& \cellcolor{orange!40}\textbf{5.112}
& \cellcolor{orange!40}\textbf{0.4843}
& \cellcolor{orange!40}\textbf{0.5873}
& \cellcolor{orange!40}\textbf{0.6435} \\
Stochastic DMD 
& \cellcolor{orange!40}\textbf{1.019}
& \cellcolor{orange!40}\textbf{32.86}
& \cellcolor{orange!40}\textbf{28.85}
& \cellcolor{orange!20}2.572
& \cellcolor{orange!20}0.2213
& \cellcolor{orange!20}0.2594
& \cellcolor{orange!20}0.4795 \\
GAN only
& \cellcolor{orange!20}0.849
& \cellcolor{orange!20}32.52
& \cellcolor{orange!20}34.19
& 2.219 & 0.1922 & 0.2080 & 0.4007 \\
\bottomrule
\end{tabular}
\caption{\textbf{Ablation of PMD components.} All variants share the same stochastic student architecture. PMD only removes the GAN loss, while GAN only removes the PMD loss. Dark orange indicates the best result and light orange the second-best result for each metric.}
\label{tab:pmd_components}
\end{table*}

\paragraph{FID.}
FID~\cite{heusel2017gans} compares the distributions of Inception-V3 features from the 10,000 generated images and their corresponding real COCO images:
\begin{equation}
  \begin{split}
    \operatorname{FID}
    &=\|\mu_g-\mu_r\|_2^2\\
    &\quad+\operatorname{Tr}\!\left(
       \Sigma_g+\Sigma_r-2(\Sigma_g\Sigma_r)^{1/2}\right),
  \end{split}
  \label{eq:app_metric_fid}
\end{equation}
where $\mu$ and $\Sigma$ are feature means and covariances. We use clean-FID with identical center-cropping and resizing for real and generated images. FID is computed over the complete image sets; lower is better.

\section{Analysis on Tilt $\lambda$}
\label{sec:analysis_tilt_direction}

\subsection{Tilt Direction}
\label{subsec:tilt_direction}

We investigate isotropic and spatially coarse tilt directions to
understand how their spatial structure affects posterior matching, as illustrated in Fig.~\ref{app:tilt_selection_ablation}.
The Tilt=Shift Identity provides a geometric perspective on why
coarse directions might yield greater sample diversity than
isotropic ones. Recall the forward noising process
$z_t = \alpha_t x + \sigma_t \epsilon$.

\begin{pmdkey}
\noindent\textbf{Tilt as a Clean-Signal Shift.}
Let $\rho_t = \sigma_t^2/\alpha_t$. Applying a tilt to the
noisy input gives
\begin{equation}
\begin{aligned}
z_t + \rho_t \lambda
&= \alpha_t x + \sigma_t \epsilon + \rho_t \lambda \\
&= \alpha_t \left(x + \frac{\rho_t}{\alpha_t}\lambda\right)
   + \sigma_t \epsilon.
\end{aligned}
\label{eq:tilt_clean_shift}
\end{equation}
\end{pmdkey}

This shows that the shifted input can be interpreted as the same
Gaussian noising process applied to a displaced clean sample.
The tilt leaves the noise realization and noise level unchanged,
while shifting the underlying clean signal.

Natural images exhibit strong spatial correlations and occupy
a relatively low-dimensional region of the high-dimensional
ambient space. Isotropic tilts perturb neighboring positions
independently, whereas coarse tilts introduce spatially coherent
changes that better reflect the structure of natural image
variations. Such directions may move the clean signal toward
other plausible image configurations or nearby modes of the
conditional posterior.

This perspective offers an intuition for our empirical observation
that coarse tilts yield greater diversity: with a limited number
of teacher evaluations, spatially coherent directions may probe
more meaningful variations in the clean-image distribution,
providing richer supervision for learning the teacher's
conditional posterior.

\subsection{Tilt Norm}
\label{subsec:tilt_norm}

We ablate different tilt radii to understand how the strength of
exponential tilting affects posterior exploration and sample diversity.
For small $\|\lambda\|$, the exponentially tilted mean admits
the first-order expansion
\begin{equation}
\frac{\mathbb{E}[\hat{x}e^{\lambda^\top \hat{x}}]}
     {\mathbb{E}[e^{\lambda^\top \hat{x}}]}
=
\mathbb{E}[\hat{x}]
+
\mathrm{Cov}(\hat{x})\lambda
+
O(\|\lambda\|^2).
\end{equation}
As $\|\lambda\| \to 0$, the objective approaches ordinary mean
matching, with limited sensitivity to covariance and higher-order
posterior structure. Increasing the tilt radius strengthens the
exponential reweighting, allowing PMD to probe more pronounced
variations and emphasize alternative modes of the conditional
posterior. However, excessively large tilts can concentrate the
exponential weights on a few student samples, making finite-sample
estimation less stable.

As shown in Table~\ref{tab:tilt_radius_ablation}, larger tilt radii consistently improve
diversity, with $\mathcal{U}(1,2)$ achieving the highest scores
across all four diversity metrics. In contrast, sampling radii
from a broader range, $\mathcal{U}(0.25,2)$, achieves the best
FID. These results suggest a trade-off between exploring diverse
posterior modes and maintaining generation quality, motivating
the use of a mixture of smaller and larger tilt magnitudes.

\begin{figure*}[t]
    \centering
    \includegraphics[width=1.0\linewidth]{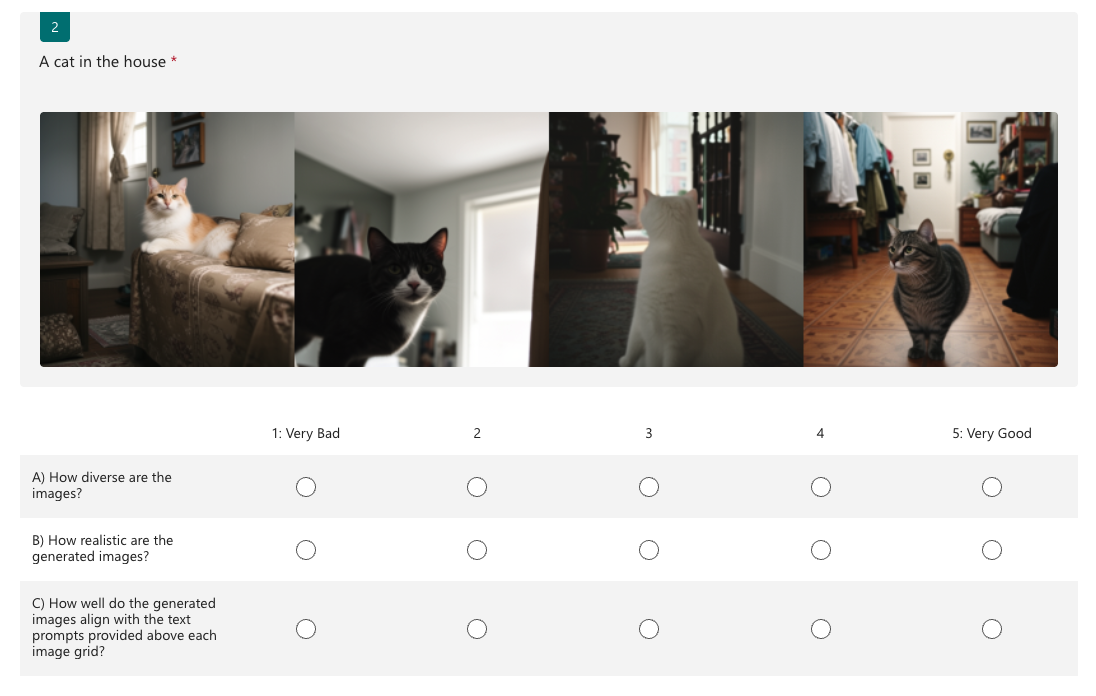} \\
    \caption{\textbf{User Study Form.}}
    \label{fig:user_study_form}
\end{figure*}

\section{More Ablation Studies}
\label{app:more_ablations}
\subsection{Ablation on PMD Components}
\label{app:pmd_components}

Table~\ref{tab:pmd_components} isolates the two terms of the student objective
$\mathcal{L}_{\text{student}} = \mathcal{L}_{\text{PMD}} + \gamma_{\text{GAN}}\mathcal{L}_{\text{GAN}}$
in the four-step FLUX.2 [klein] 4B setting.
All variants use the same stochastic student architecture, so the differences are attributable to the training objective rather than to the noise input.

\noindent  \textbf{PMD only.}
Removing the GAN loss yields the highest intra-prompt diversity on all four metrics, with a Vendi score of 5.112 and DreamSim, DINOv2, and LPIPS diversities of 0.4843, 0.5873, and 0.6435.
Relative to Stochastic DMD, this is roughly $2.0\times$, $2.2\times$, and $2.3\times$ higher on Vendi, DreamSim, and DINOv2, respectively.
The same variant has the weakest quality and alignment scores (ImageReward $-1.232$, CLIP-S 26.88, FID 56.13).
This behavior is expected and follows from the target that PMD matches, rather than from an optimization failure.
The Tilt=Shift Identity and the Posterior Identification Theorem assume an exact posterior-mean denoiser, so all PMD targets are evaluations of the frozen teacher at guidance scale 1; a classifier-free-guided prediction is, in general, not the posterior mean of any fixed clean-image distribution under Gaussian corruption.
The distribution identified at zero PMD loss is therefore the teacher's \emph{unguided} conditional posterior.
Text-to-image teachers sampled without guidance are known to produce broad but low-fidelity and weakly prompt-aligned samples, and a student that matches this posterior faithfully inherits both properties.
The low ImageReward and CLIP-S of the PMD-only student thus reflect the unguided teacher distribution, and its diversity scores should be read together with these quality scores, since part of the measured variation comes from samples that the guided teacher would not produce.

\noindent  \textbf{GAN only.}
Removing the PMD loss gives reasonable quality (ImageReward 0.849, CLIP-S 32.52, FID 34.19) but the lowest diversity on every metric (Vendi 2.219, DreamSim 0.1922, DINOv2 0.2080, LPIPS 0.4007).
Although the student still receives an independent noise input, the discriminator scores each sample individually and provides no incentive to use that input to cover the posterior.
The diversity of PMD therefore comes from the posterior-matching objective and not from the stochastic architecture alone.

\noindent  \textbf{Stochastic DMD.}
Training the same stochastic student with the DMD objective gives the best quality (ImageReward 1.019, CLIP-S 32.86, FID 28.85), but its diversity is only marginally above the GAN-only variant (Vendi 2.572 vs.\ 2.219) and about half that of the PMD-only variant.
This is consistent with the mode-seeking behavior of the reverse-KL objective: making the student stochastic is not sufficient, and the objective itself must reward coverage of the conditional posterior.
\begin{figure*}[t]
    \centering
    \includegraphics[width=1.0\linewidth]{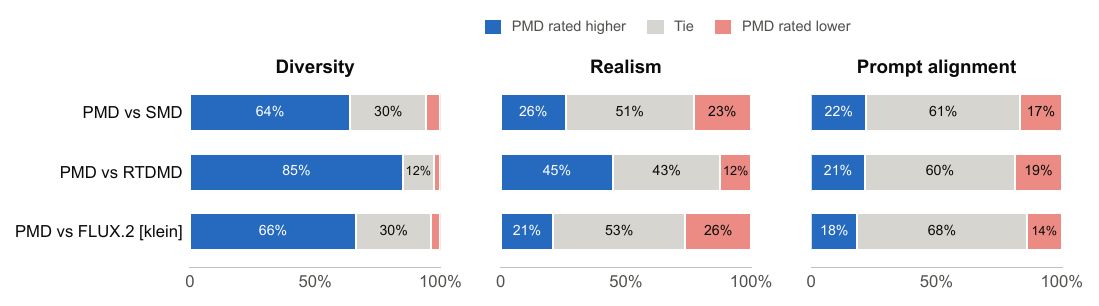} \\
    \caption{\textbf{User Study for FLUX.2 [klein] 4B four-step models.}}
    \label{fig:full_user_study}
\end{figure*}
\noindent \textbf{Discussion.}
The two terms of the PMD objective are complementary.
The PMD loss supplies coverage of the teacher's conditional posterior, while the GAN loss, which compares student samples with real images under caption conditioning, supplies the fidelity and prompt alignment that the unguided posterior lacks.
PMD on its own is therefore not intended as a complete training objective: it requires an auxiliary loss to reach competitive visual quality, and we use the adversarial term for this purpose.
Their combination gives the full four-step FLUX.2 [klein] 4B model reported in the main paper, whose diversity lies between the two extremes of this ablation (Vendi 3.371, compared with 5.112 for PMD only and 2.572 for Stochastic DMD) while reaching competitive quality at guidance scale 1.
The adversarial term thus trades part of the unguided posterior's breadth for realism, and the resulting student remains substantially more diverse than the DMD-trained one.

\section{User Study Details}
\label{app:user_study}

We conducted a human evaluation with 50 participants to assess the perceptual quality of Posterior Matching Distillation (PMD) against three baselines: SMD, RTDMD, and FLUX.2 [klein]. The evaluation considered three criteria: \textbf{diversity}, \textbf{realism}, and \textbf{prompt alignment}.

\paragraph{Evaluation Protocol.}
We selected 24 text prompts and generated four images per prompt using different random seeds, resulting in 96 images per method. For each prompt, the four generated images were presented together in an image grid, allowing participants to assess variation across samples conditioned on the same text description as shown in~\ref{fig:user_study_form}. Participants were asked to rate each image grid along three dimensions:
\begin{enumerate}
    \item \textbf{Diversity:} How diverse are the generated images?
    \item \textbf{Realism:} How realistic are the generated images?
    \item \textbf{Prompt alignment:} How well do the generated images align with the provided text prompt?
\end{enumerate}
Each criterion was evaluated on a five-point Likert scale, ranging from 1 (\emph{Very Bad}) to 5 (\emph{Very Good}). For each baseline, we compared its ratings with those of PMD and computed the percentage of comparisons in which PMD received a higher rating, an equal rating (tie), or a lower rating. The aggregated results are presented in Fig.~\ref{fig:full_user_study}.

\paragraph{Results.}
PMD substantially improves sample diversity, receiving higher ratings in
\textbf{64\%, 85\%, and 66\%} of comparisons against SMD, RTDMD, and
FLUX.2 [klein], respectively, while receiving lower ratings in only
6\%, 3\%, and 4\%. For realism and prompt alignment, most comparisons
result in ties, indicating broadly comparable performance. PMD shows
its strongest realism advantage over RTDMD, winning 45\% of comparisons
versus 12\% losses. Overall, these results demonstrate that PMD
significantly enhances diversity while maintaining competitive
realism and prompt adherence.

\begin{figure*}[t]
    \centering
    \includegraphics[width=1.0\linewidth]{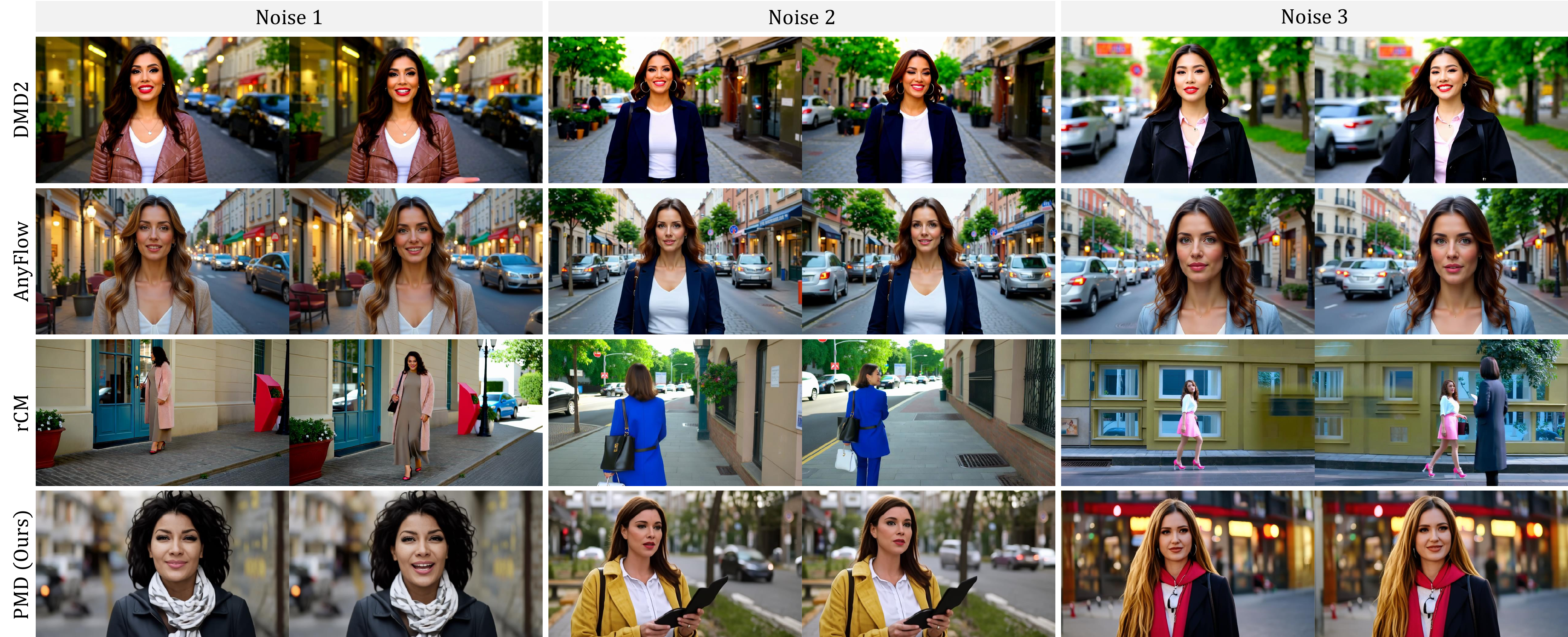} \\
    \caption{\textbf{Diversity and visual quality comparison on Wan 2.1 1.3B.} Four-step generation results for the prompt \textit{“A woman in a street”} across three noise seeds. DMD2 and AnyFlow produce visually similar facial structures and appearances across different seeds, indicating limited generative diversity, while rCM exhibits substantial degradation in visual quality. In contrast, our Posterior Matching Distillation (PMD) generates diverse facial identities, poses, clothing, and street backgrounds while maintaining high visual fidelity and prompt alignment. These results highlight PMD's ability to preserve sample diversity and generation quality under the same four-step sampling budget.}
    \label{fig:wan_comparison}
\end{figure*}

\begin{table*}[t]
\centering
\small
\setlength{\tabcolsep}{5pt}
\resizebox{\textwidth}{!}{%
\begin{tabular}{@{}lcccccccc@{}}
\toprule
& &
\multicolumn{3}{c}{VBench} &
\multicolumn{2}{c}{V-JEPA 2} &
\multicolumn{2}{c}{VideoMAE V2} \\
\cmidrule(lr){3-5}
\cmidrule(lr){6-7}
\cmidrule(lr){8-9}
Method & NFE &
Overall$\uparrow$ & Quality$\uparrow$ & Semantic$\uparrow$ &
Cosine$\uparrow$ & L2$\uparrow$ &
Cosine$\uparrow$ & L2$\uparrow$ \\
\midrule

\textcolor{gray}{UniPC (Teacher)}
& \textcolor{gray}{$50\times2$}
& \textcolor{gray}{84.18}
& \textcolor{gray}{85.33}
& \textcolor{gray}{79.56}
& \textcolor{gray}{0.1276}
& \textcolor{gray}{27.65}
& \textcolor{gray}{0.02832}
& \textcolor{gray}{2.928} \\
\midrule

rCM~\cite{zheng2026large}
& 4
& 83.28 & 82.25 & 78.16
& \cellcolor{orange!40}\textbf{0.1050}
& \cellcolor{orange!40}\textbf{25.18}
& 0.02036
& \cellcolor{orange!20}2.671 \\

AnyFlow~\cite{gu2026anyflow}
& 4
& \cellcolor{orange!40}\textbf{85.51}
& \cellcolor{orange!20}86.41
& \cellcolor{orange!20}81.91
& 0.0795 & 21.19 & 0.01076 & 2.401 \\

DMD2~\cite{yin2024improved}
& 4
& 84.50 & 85.81 & 79.26
& 0.0894 & 22.51 & 0.01601 & 2.193 \\

PDD (Midpoint)$^{\dagger}$~\cite{shaul2026parallel}
& 4
& 84.94
& \cellcolor{orange!40}\textbf{86.45}
& 78.91
& 0.1032 & 24.63
& \cellcolor{orange!20}0.02054
& 2.548 \\

PMD (Ours)
& 4
& \cellcolor{orange!20}85.47
& 86.33
& \cellcolor{orange!40}\textbf{82.04}
& \cellcolor{orange!20}0.1038
& \cellcolor{orange!20}25.03
& \cellcolor{orange!40}\textbf{0.02164}
& \cellcolor{orange!40}\textbf{2.692} \\

\bottomrule
\end{tabular}%
}
\vspace{3pt}

\begin{minipage}{\textwidth}
\footnotesize
$^{\dagger}$ PDD (Midpoint) results are taken directly from
the reported PDD results rather than independently evaluated
on our Self-Forcing prompt set.
\end{minipage}

\caption{\textbf{Wan 2.1 1.3B text-to-video generation results.}
Performance and diversity on VBench using the MovieGenVideoBench
prompt set. Diversity is measured using pairwise V-JEPA 2 and
VideoMAE V2 feature distances across five generated videos per
prompt. Externally reported results are marked with $\dagger$.
Dark orange indicates the best result and light orange the
second-best result among four-step methods.}
\label{tab:wan_video}
\end{table*}

\section{FLUX.2 [klein] 4B Results}
\label{sec:Flux_results}

\begin{figure*}[p]
    \centering
    \includegraphics[width=1.0\linewidth]{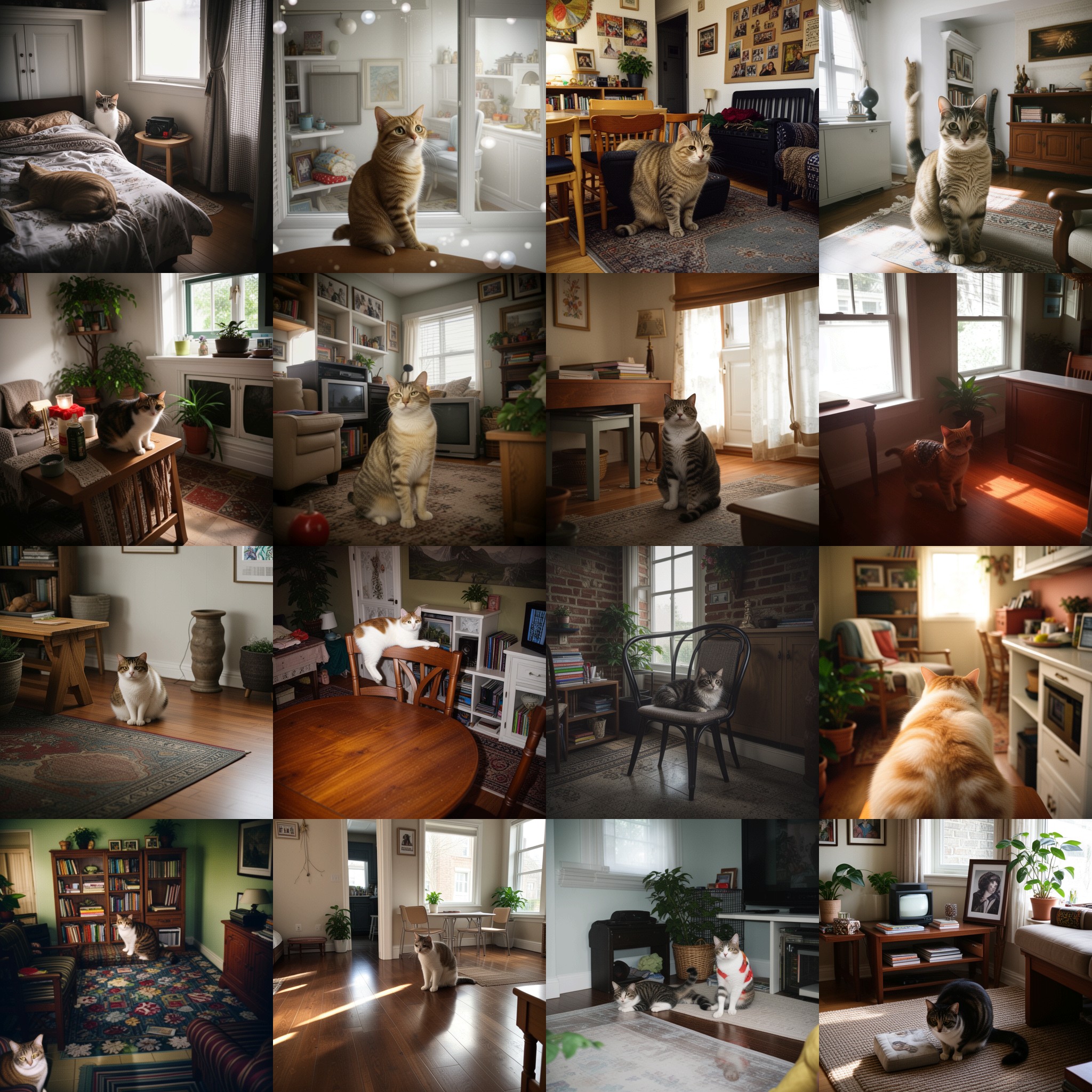} \\
    \caption{FLUX.2 [klein] 4B 4-step generation results for the prompt "A cat in the house" with seeds 50-65.}
    \label{fig:flux_appendix_cat}
\end{figure*}

\begin{figure*}[t]
    \centering
    \includegraphics[width=1.0\linewidth]{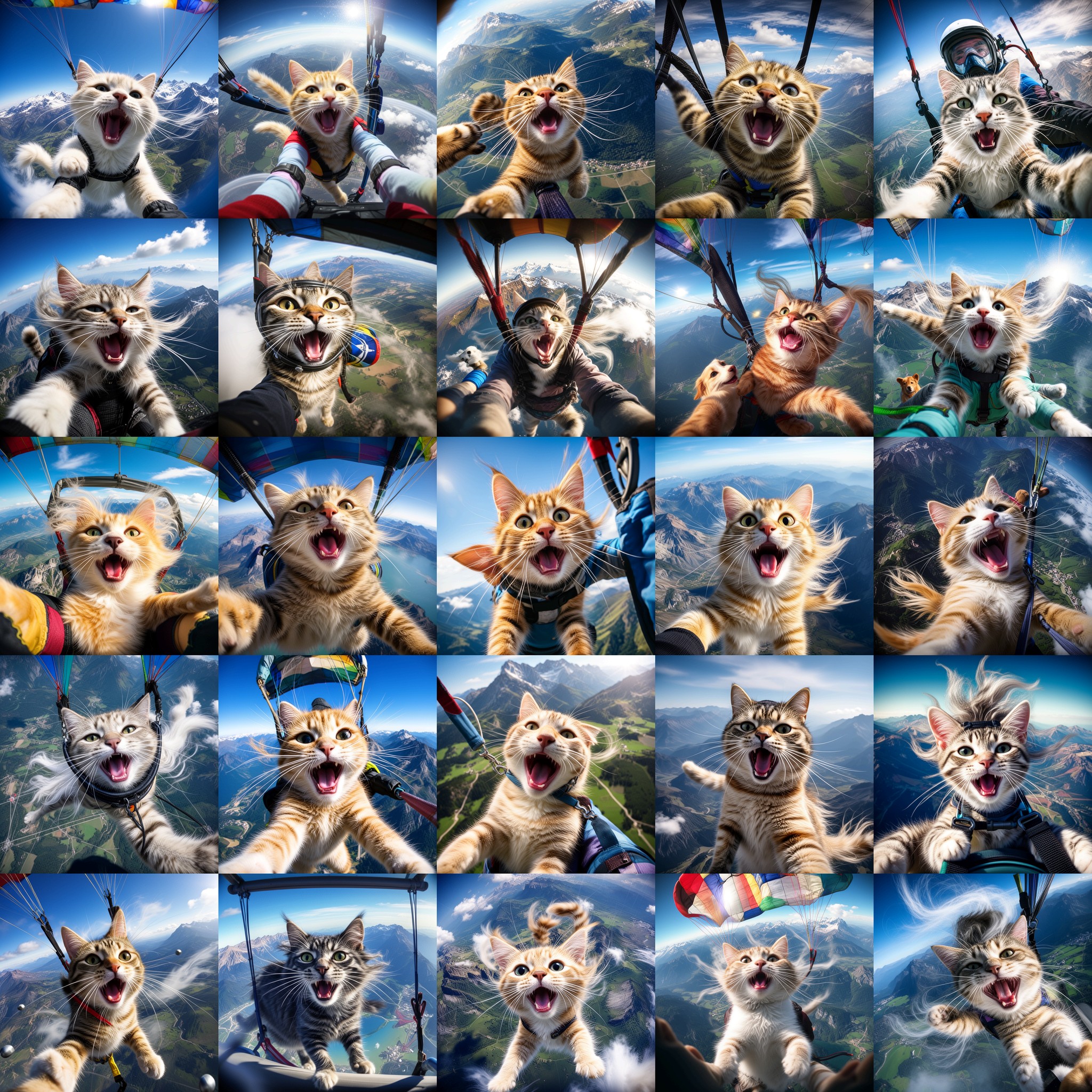} \\
    \caption{FLUX.2 [klein] 4B 4-step generation results for the prompt "Surreal POV selfie shot from a cute baby cat with a parachute during a skydiving adventure. The breathtaking aerial perspective shows a alpine landscape. The cute baby cat fur ripples chaotically in the extreme wind, and its wide-open mouth reveals an expression of both exhilaration and absurdity. its lips and cheeks flutter wildly as the intense winds distort its face in humorous, exaggerated ways." with seeds 50-74.}

    \label{fig:flux_appendix_cat_skydive}
\end{figure*}

We share qualitative results of our 4-step PMD model on FLUX.2 [klein] 4B in Fig.~\ref{fig:flux_appendix_cat} to Fig.~\ref{fig:flux_appendix_cat_skydive}.

\begin{figure*}[t]
    \centering
    \includegraphics[width=1.0\linewidth]{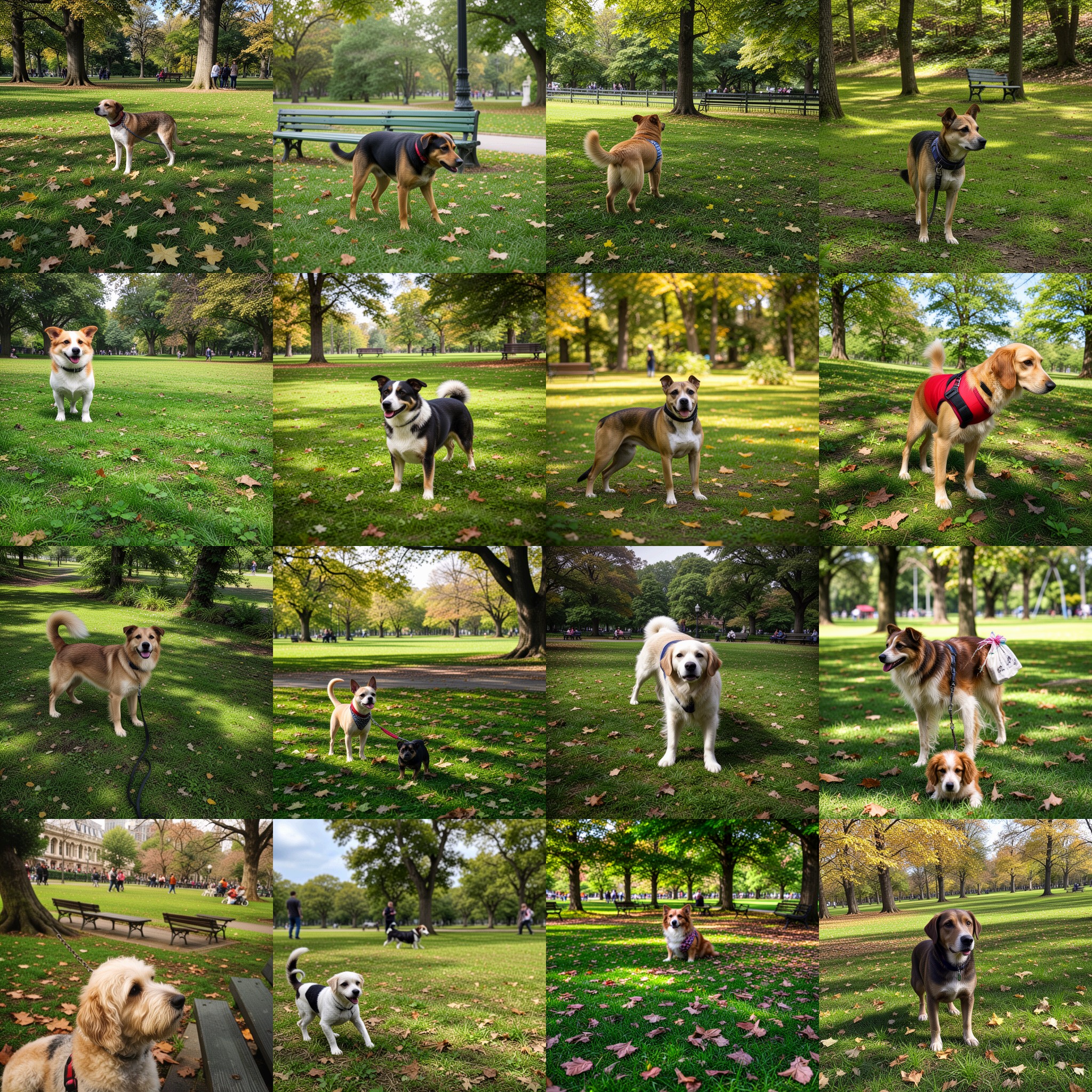} \\
    \caption{FLUX.2 [klein] 4B 4-step generation results for the prompt "A dog in the park" with seeds 50-65.}
    
\end{figure*}

\begin{figure*}[t]
    \centering
    \includegraphics[width=1.0\linewidth]{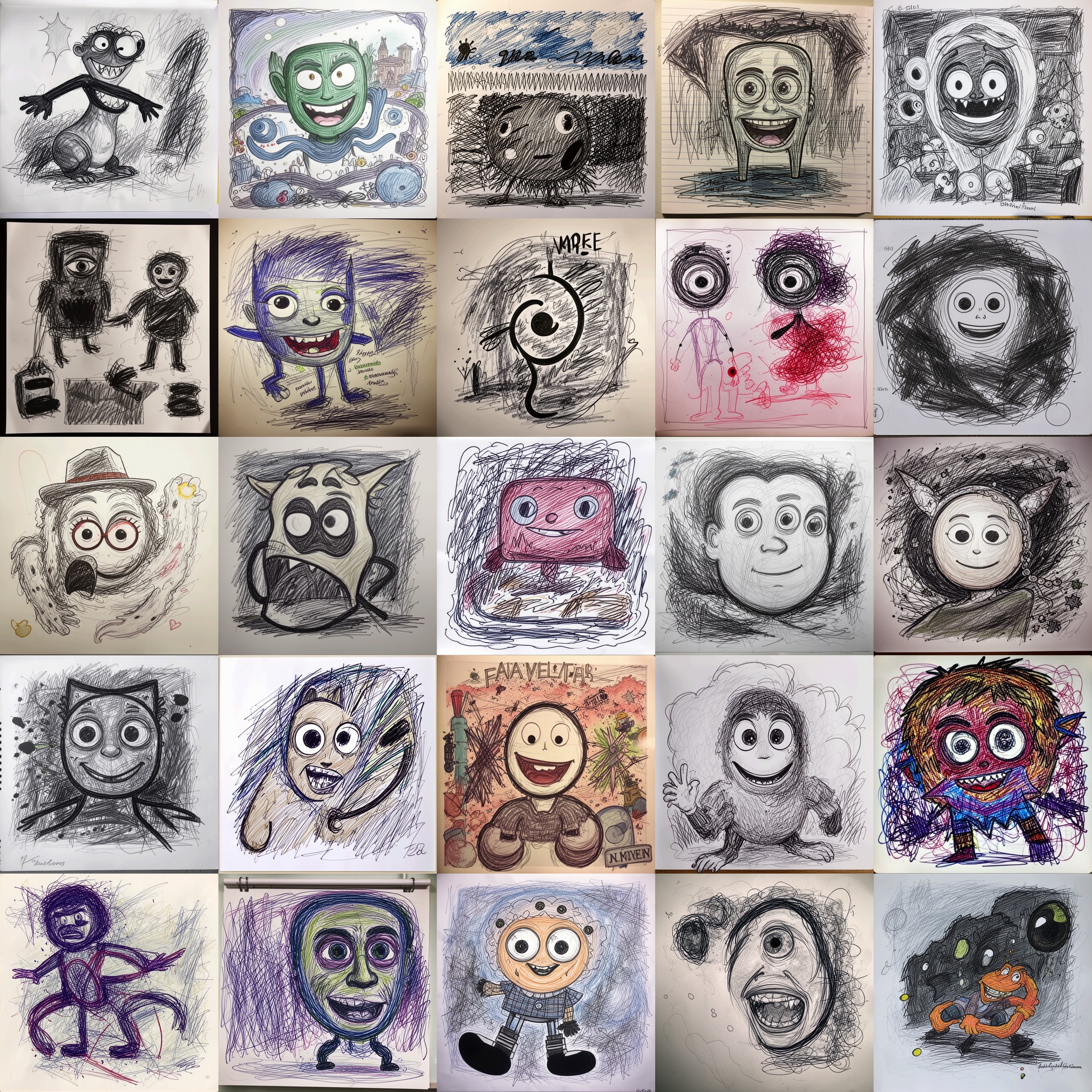} \\
    \caption{FLUX.2 [klein] 4B 4-step generation results for the prompt "Scary chaotic scribbles of a nightmarish drawing of Mike Wazowski, done by a 3 year old on white construction paper" with seeds 50-74.}
    
\end{figure*}

\begin{figure*}[t]
    \centering
    \includegraphics[width=1.0\linewidth]{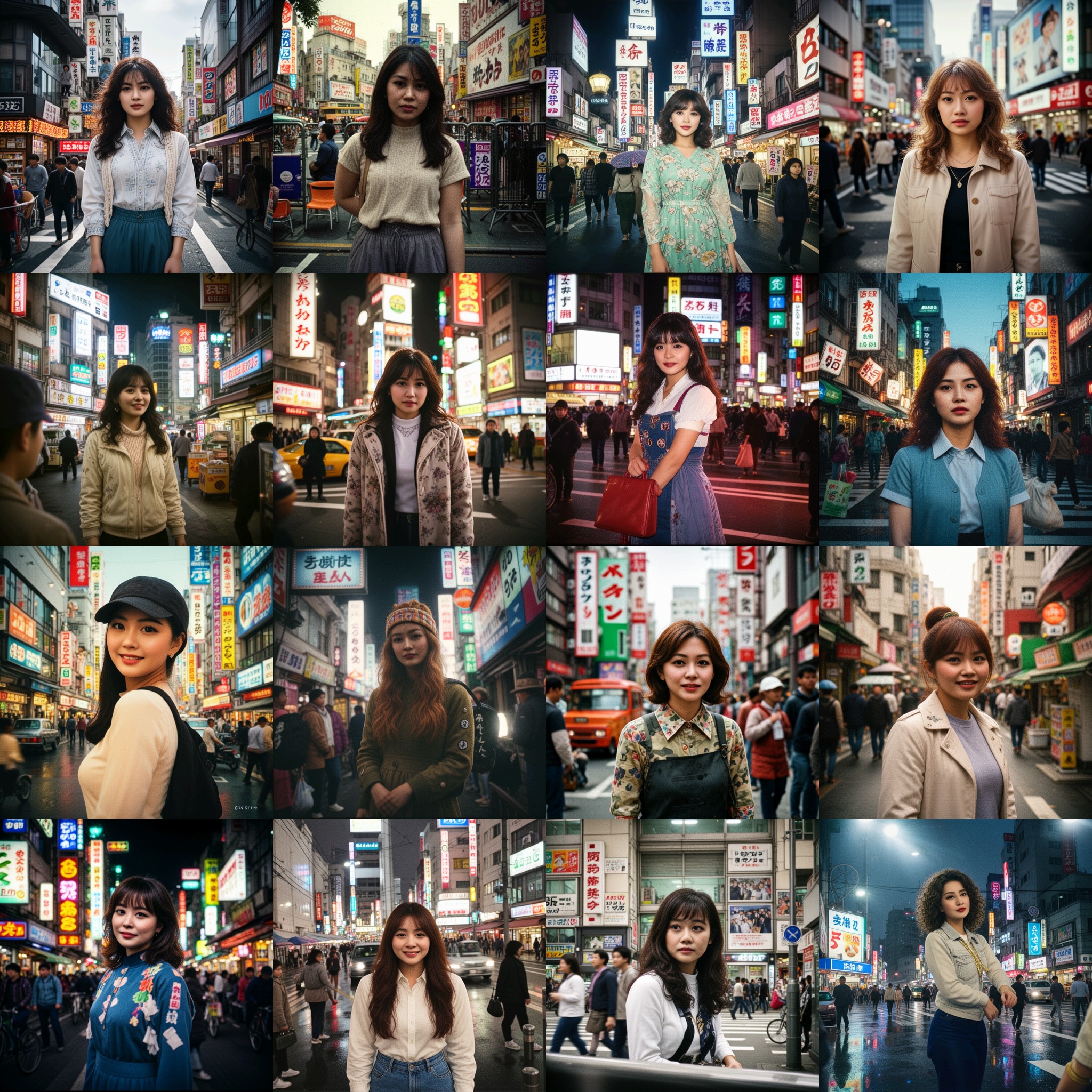} \\
    \caption{FLUX.2 [klein] 4B 4-step generation results for the prompt "A pretty girl in Tokyo, 80's" with seeds 50-65.}
    
\end{figure*}

\begin{figure*}[t]
    \centering
    \includegraphics[width=1.0\linewidth]{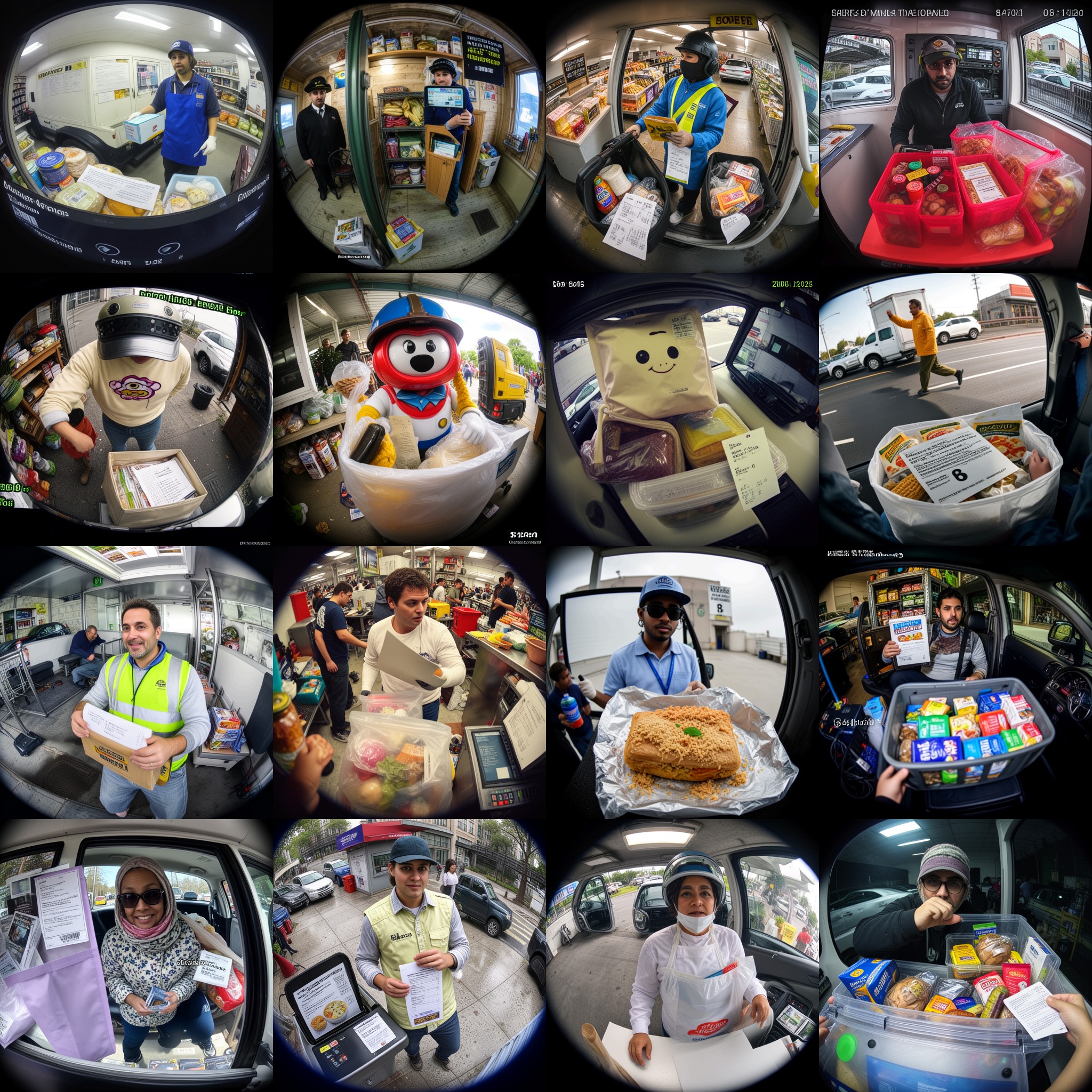} \\
    \caption{FLUX.2 [klein] 4B 4-step generation results for the prompt "Ring camera footage of a food delivery driver" with seeds 50-65.}
    
\end{figure*}

\section{Video Generation Results}
\label{sec:video_results}

We evaluate PMD on Wan 2.1 1.3B using four-step text-to-video generation.
Table~\ref{tab:wan_video} compares PMD against rCM, AnyFlow, DMD2,
and PDD in terms of generation quality, semantic consistency, and
sample diversity. We measure video quality using VBench and evaluate
diversity through pairwise feature distances from V-JEPA 2 and
VideoMAE V2 across five generated videos per prompt.

\noindent\textbf{Generation Quality.}
PMD achieves the highest VBench Semantic score of 82.04 and a
competitive Quality score of 86.33, resulting in an Overall score
of 85.47, closely matching AnyFlow's 85.51. Notably, PMD
surpasses the 50-step teacher reference in all three VBench
metrics while requiring only four network evaluations.
These results demonstrate that PMD preserves semantic consistency
and visual quality under a substantially reduced sampling budget.

\noindent\textbf{Generation Diversity.}
PMD achieves the highest VideoMAE V2 diversity scores, with
0.02164 cosine distance and 2.692 L2 distance, while ranking
second on both V-JEPA 2 metrics. Interestingly, rCM also
demonstrates strong diversity, achieving the highest V-JEPA 2
scores of 0.1050 and 25.18. However, our qualitative observations
reveal an important distinction: despite producing diverse
video content, rCM outputs often appear overly smooth, lacking
high-frequency details and fine textures. This is also reflected
in its comparatively lower VBench Quality score. In contrast,
PMD generates visually detailed videos while maintaining strong
variation across samples. These findings highlight the importance
of evaluating diversity alongside perceptual quality, as high
feature-space diversity alone does not necessarily indicate
high-fidelity video generation.

\end{document}